%% file: Archive_HI-SAFE.tex
\documentclass[journal]{IEEEtran}
\input{macros}

\usepackage[final]{graphicx}
\usepackage{amsmath,amsfonts}
\usepackage{amsthm}
\usepackage{algorithmic}
\usepackage{algorithm}
\usepackage{array}
\usepackage{caption}
\usepackage{subcaption}
\usepackage{textcomp}
\usepackage{stfloats}
\usepackage{url}
\usepackage{verbatim}
\usepackage{graphicx}
\usepackage{cite}
\usepackage{color}
\usepackage{amssymb}

\allowdisplaybreaks
\usepackage{multirow}
\newtheorem{theorem}{Theorem}
\newtheorem*{theorem*}{Theorem}
\newtheorem{corollary}{Corollary}[theorem]
\newtheorem{lemma}{Lemma}
\newtheorem{remark}{Remark}

\usepackage{stmaryrd}
\usepackage{kotex}
\usepackage{booktabs}
\usepackage{pifont}
\usepackage{enumitem}

\ifCLASSINFOpdf
\else
\fi
\begin{document}
\title{Hi-SAFE: Hierarchical Secure Aggregation for Lightweight Federated Learning}

\author{Hyeong-Gun Joo,~\IEEEmembership{Member,~IEEE,} Songnam Hong,~\IEEEmembership{Member,~IEEE,} Seunghwan Lee,  and Dong-Joon Shin,~\IEEEmembership{Senior,~IEEE}% <-this % stops a space
\thanks{The authors are with the Department of Electronic Engineering, Hanyang University, Seoul, South Korea, Corresponding author: Dong-Joon Shin, e-mail: djshin@hanyang.ac.kr.}% <-this % stops a space
%\thanks{This paper was produced by the IEEE Publication Technology Group. They are in Piscataway, NJ.}% <-this % stops a space
}

%\thanks{Manuscript received April 19, 2024; revised August 16, 2024.}}

%\markboth{Journal of \LaTeX\ Class Files,~Vol.~14, No.~8, April~2024}%
%{Joo \MakeLowercase{\textit{et al.}}: A Sample Article Using IEEEtran.cls for IEEE Journals}


\maketitle

\begin{abstract}
Federated learning (FL) faces challenges in ensuring both privacy and communication efficiency, particularly in resource-constrained environments such as Internet of Things (IoT) and edge networks. While sign-based methods, such as sign stochastic gradient descent with majority voting (\textsc{signSGD-MV}), offer substantial bandwidth savings, they remain vulnerable to inference attacks due to exposure of gradient signs. Existing secure aggregation techniques are either incompatible with sign-based methods or incur prohibitive overhead. 
To address these limitations, we propose \emph{Hi-SAFE}, a lightweight and cryptographically secure aggregation framework for sign-based FL. Our core contribution is the construction of efficient majority vote polynomials for \textsc{signSGD-MV}, derived from Fermat’s Little Theorem. This formulation represents the majority vote as a low-degree polynomial over a finite field, enabling secure evaluation that hides intermediate values and reveals only the final result. We further introduce a hierarchical subgrouping strategy that ensures constant multiplicative depth and bounded per-user complexity, independent of the number of users $n$. 
Hi-SAFE reduces per-user communication by over 94\% when $n \geq 24$, and total cost by up to 52\% at $n = 24$, while preserving model accuracy. Experiments on benchmark datasets confirm the scalability, robustness, and practicality of Hi-SAFE in bandwidth-constrained FL deployments.
\end{abstract}

\begin{IEEEkeywords}
Communication efficiency, federated learning, privacy-preserving, secure aggregation, subgrouping.
\end{IEEEkeywords}

\IEEEpeerreviewmaketitle

\section{Introduction}
\label{sec:intro}
\IEEEPARstart{F}{ederated} learning (FL) facilitates collaborative model training across decentralized clients without exposing raw data~\cite{Mcmahan2017,ghimire2022recent, rauniyar2023federated,Li2020survey,hong2021communication,kwon2023tighter,lim2020federated,Yang2023VFL}, offering intrinsic privacy benefits that make it particularly attractive for sensitive domains such as healthcare, finance, and the Internet of Things (IoT). Nonetheless, deploying FL on real-world edge or IoT devices introduces significant challenges due to limited communication bandwidth, computational capacity, and vulnerability to privacy leakage through shared model updates~\cite{Lyu2022,Nguyen2021,Aledhari2022,kairouz2021}.
Although FL effectively preserves data locality, numerous studies have shown that intermediate model updates, such as gradients, can be exploited by adversaries to reconstruct sensitive inputs or perform membership inference~\cite{Zhu2019,Hitaj2017,Geiping2020,Nasr2019,Wei2021}. These threats are especially pronounced in resource-constrained environments where devices continuously collect and transmit private information.

To address this, various secure aggregation methods have been proposed. %Pairwise additive masking~\cite{Bonawitz2017, So2022} cryptographically conceals individual updates; however, it may inadvertently leak intermediate summations. 
Pairwise additive masking~\cite{Bonawitz2017, So2022} protects individual updates via secret sharing but may still expose intermediate aggregation results under semi-honest assumptions.
Differential privacy (DP)~\cite{Truex2019a,lyu2021dp} provides formal privacy guarantees but often compromises model accuracy due to added noise. %Homomorphic encryption (HE)~\cite{cheon2017,Fang2021} offers strong cryptographic protection, enabling computations on encrypted data without decryption. However, this approach incurs substantial computational and communication costs, severely limiting its practicality on edge devices.
Homomorphic encryption (HE)~\cite{cheon2017,Fang2021} provides strong cryptographic guarantees by enabling computations directly on encrypted data without decryption. However, this approach entails substantial computational and communication costs, which significantly limits its practicality in resource-constrained edge devices.

In parallel, sign-based methods such as \textsc{signSGD} and its majority vote variant \textsc{signSGD-MV}~\cite{Sei14,Ber18, Ber19,park2023,jin2024sign,Joo2025} provide exceptional communication efficiency by quantizing updates to 1 bit per parameter. These methods are both scalable and robust to noise; however, they expose raw sign gradients to the server, rendering them susceptible to inference attacks~\cite{Geiping2020}. Moreover, most existing secure aggregation protocols are either inefficient or fundamentally incompatible with sign-based methods. Specifically, masking-based approaches permit the server to access intermediate summation values during the computation of the final majority vote, which may lead to information leakage. HE cannot directly support nonlinear operations—such as the sign function and majority voting—required by \textsc{signSGD-MV}. Additionally, the large ciphertext sizes in HE undermine the key benefit of 1-bit update protocols. %A comprehensive comparative summary of these approaches is presented in Appendix~\ref{appendix:existing_secureAGG}, highlighting their limitations in the context of sign-based FL.

\vspace{2mm}
\textit{These limitations motivate the necessity for a novel class of secure aggregation frameworks that not only preserve the communication efficiency characteristic of sign-based methods but also provide strong privacy guarantees.}

% non-subgrouping은 "by revealing only the final majority vote" 가능하고, subgrouping은 subgroup의 MV 볼수 있음.
\subsection{Contributions}  
\begin{figure}[t]
\centering
\includegraphics[width=0.48\textwidth]{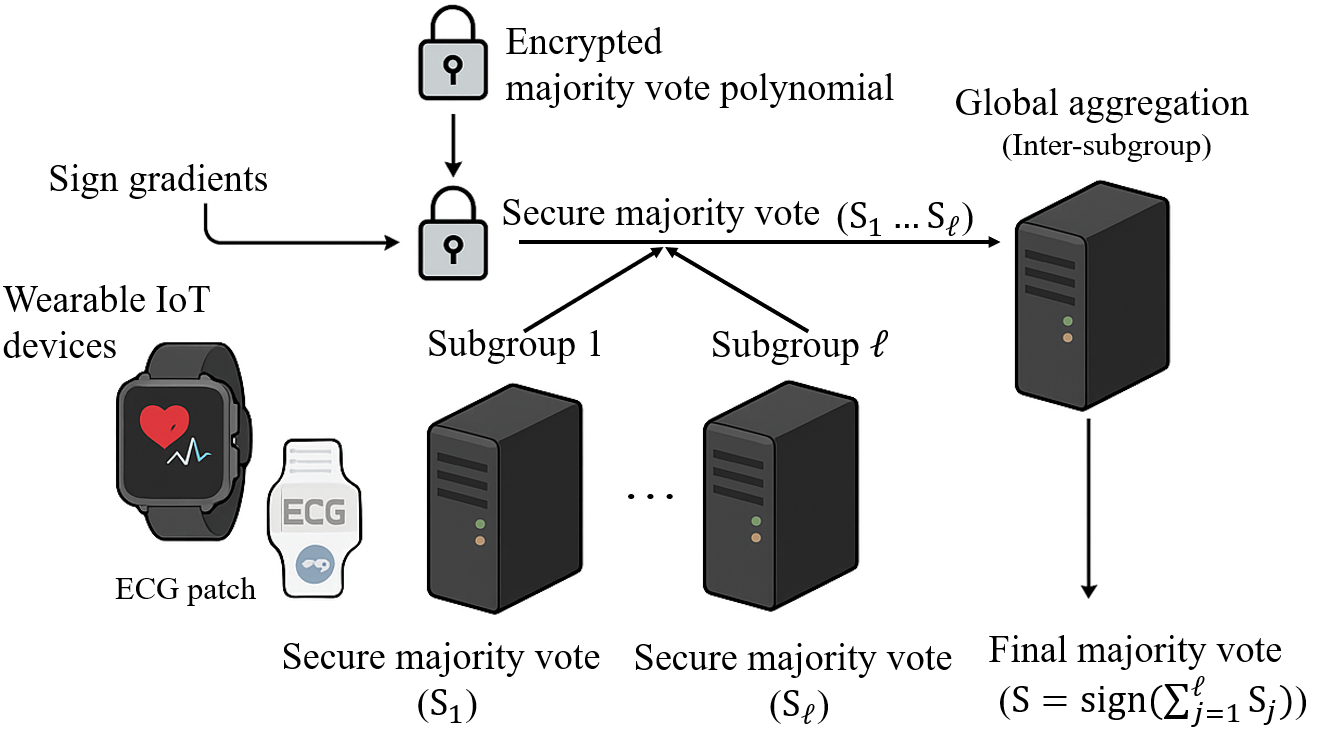} %0.77
\caption{\textbf{Hi-SAFE}: Hierarchical Secure Aggregation Framework.}
%\caption{Overview of \textbf{Hi-SAFE}. %Sign-based local gradients are securely aggregated in a two-level hierarchy through polynomial evaluation. Each user contributes its 1-bit gradient for privacy-preserving intra-subgroup aggregation, and only the majority vote result is revealed to the server, ensuring end-to-end privacy without exposing individual inputs.}
\label{fig:hisafe_overview}
\end{figure}

%\noindent\textbf{Contributions.}  
To address this challenge, we propose \emph{\textbf{Hi-SAFE}} (\textbf{Hi}erarchical \textbf{S}ecure \textbf{A}ggregation for \textbf{FE}derated Learning)—a lightweight and cryptographically secure aggregation framework tailored to \textsc{signSGD-MV}. Hi-SAFE minimizes communication cost, protects against inference attacks by revealing only the majority vote result, and scales efficiently in resource-constrained environments. As illustrated in Fig.~\ref{fig:hisafe_overview}, each user contributes a 1-bit signed update that is securely processed through evaluation of majority vote polynomial in a hierarchical structure. Our main contributions are summarized as follows: 
\begin{itemize}[leftmargin=0.4em]
    \item \textbf{Cryptographic Secure Aggregation:} We design a privacy-preserving protocol for sign-based FL that discloses only the final majority vote to the server, thereby ensuring protection against inference attacks under the semi-honest model. To the best of our knowledge, this is the first work to provide end-to-end privacy within sign-based FL frameworks.

    \item \textbf{Efficient Majority Vote Polynomial:} Based on Fermat’s Little Theorem, we construct the majority vote as a low-degree polynomial over a finite field and show that its secure evaluation is equivalent to the standard (non-private) \textsc{signSGD-MV}, guaranteeing both correctness and privacy.

    \item \textbf{Hierarchical Scalability:} We introduce a subgrouping strategy that maintains constant multiplicative depth (about two subrounds) and a bounded secure multiplication cost ($\leq 6$ per user), independent of the total number of users $n$.

    \item \textbf{Communication-Efficient and Robust Framework:} Hi-SAFE reduces per-user communication costs by over 94\% when $n\! \geq \!24$, and achieves up to 52\% reduction in total communication at $n \!=\! 24$ compared with the non-subgrouping, while preserving or even improving model accuracy. Extensive experiments on benchmark datasets confirm its scalability, robustness, and practicality in bandwidth-constrained FL deployments.
\end{itemize}
 
\subsection{Organization of the Paper}
Section II introduces some previous works related to our results. A lightweight and cryptographically secure aggregation framework tailored to \textsc{signSGD-MV} is proposed in Section III. Section IV analyzes the convergence behavior and security of the proposed method. Section V evaluates the proposed methods in various environments and verifies their effectiveness in reducing communication costs through comparative analysis with the conventional method. Finally, Section VI concludes the paper.

% =====================================================================
\section{Related Work}
\label{sec:existing_secureAGG}
Numerous secure aggregation strategies have been developed to mitigate privacy risks in FL, including masking, DP, and HE. While each of these methods provides certain privacy guarantees, they exhibit significant limitations concerning communication efficiency, compatibility with sign-based protocols, and robustness against inference attacks. 

Masking-based methods~\cite{Bonawitz2017, So2022} typically employ pairwise secret sharing to cryptographically protect individual updates while ensuring correct aggregation. Although these methods are scalable, they expose intermediate aggregation results to the server or auxiliary nodes, potentially leading to information leakage under semi-honest assumptions unless additional mechanisms, such as double masking, are employed.
Local DP methods~\cite{Truex2019a, Byrd2020, lyu2021dp} perturb local model updates with noise prior to aggregation, thereby providing formal privacy guarantees. For instance, DP-SIGNSGD~\cite{lyu2021dp} adds Gaussian noise before applying the sign function. However, the presence of noisy sign gradients remains visible to the server, and achieving strong privacy often requires substantial noise, which can degrade model accuracy—especially problematic in data-sparse IoT environments.
HE~\cite{cheon2017, Zhang2020, Fang2021, Jiang2021, Ma2022, Gentry2009} allows for computation directly over encrypted data, offering strong cryptographic security. However, HE-based schemes incur significant computational overhead and produce large ciphertexts (e.g., thousands of bits per coordinate), which render them impractical for bandwidth-limited FL deployments. Additionally, HE does not support nonlinear functions, such as the sign function or majority vote, which are essential for the functionality of sign-based protocols like \textsc{signSGD-MV}.

Despite their strengths, existing secure aggregation methods are not directly compatible with sign-based protocols. Specifically, masking-based approaches permit the server to access intermediate summation values during the computation of the final majority vote, which may result in information leakage. HE-based schemes are fundamentally incompatible with nonlinear vote operations. Moreover, the high communication cost associated with HE undermines the primary advantage of \textsc{signSGD-MV}—its 1-bit update efficiency. %A detailed comparison is provided in Section~\ref{sec:existing_secureAGG1}. 

To address these limitations, we propose Hi-SAFE, a cryptographic secure aggregation framework for \textsc{signSGD-MV}. Hi-SAFE privately evaluates majority votes via secure multiplications, thereby preserving communication efficiency and enabling scalable,  privacy-preserving FL under the semi-honest model.

% --------------------------
\subsection{Comparison with Existing Secure Aggregation Methods}
\label{sec:existing_secureAGG1}

Table~\ref{tab:comparison} provides a comparative summary of the proposed Hi-SAFE framework and existing secure aggregation methods in FL. The comparison considers multiple criteria including the type of privacy guarantee, exposure level to the server, accuracy preservation, and overall communication and computational efficiency.

As summarized, existing methods such as masking and local DP offer partial protection but exhibit key limitations when applied to sign-based protocols. In particular, masking-based approaches expose intermediate summation values during majority vote computation, and DP schemes suffer from accuracy degradation due to the addition of noise. HE, while cryptographically strong, is computationally intensive and fundamentally incompatible with nonlinear operations such as $\operatorname{sign}(\cdot)$  or majority vote.

By contrast, the proposed Hi-SAFE framework achieves privacy-preserving aggregation tailored to 1-bit \textsc{signSGD-MV} by securely evaluating majority vote polynomials, instantiated for example via Beaver triples. It reveals only the final majority vote result, preserves communication efficiency, and scales well under semi-honest assumptions. 
Moreover, unlike masking-based methods that fully leak inputs in extreme cases (e.g., all users submit $-1$ or all submit $+1$), Hi-SAFE prevents such leakage by keeping all intermediate values. %Under a uniform input distribution, the probability of accidental input privacy loss is at most $1/2^{n-1}$, which is negligible in the number of users $n$.} % 현재 subgrouping 알고리즘상 중간 MV값을 볼 수 있음. 그렇지만 input을 알 수는 없음. $1/2^{n_1-1}$의 확률로 추정만 가능

\begin{table*}[!htb]
    \centering
    \renewcommand{\arraystretch}{1.7}
    \setlength{\tabcolsep}{1.7pt}
    \caption{Comparison of the proposed method with the existing privacy-preserving aggregation approaches}
    \label{tab:comparison}
    \begin{tabular}{|c|c|c|c|c|c|c|}
        \hline
        \textbf{Method} & 
        \textbf{Privacy Type} & 
        \textbf{Server Observes} & 
        \textbf{Accuracy Loss} & 
        \textbf{Comm. Efficiency} & 
        \textbf{Comp. Cost} & 
        \textbf{Scalability} \\
        \hline
        \shortstack{Masking\\\cite{Bonawitz2017}} & 
        \shortstack{Cryptographic\\(Double Masking)} & 
        \shortstack{\ding{51}\\(Summation Values)} & 
        \ding{55} & 
        Low & 
        High & 
        Limited \\
        \hline
        \shortstack{DP\\\cite{lyu2021dp}} & 
        %\shortstack{Formal\\(Local DP)} & 
        \shortstack{Statistical \\(ε-LDP)} &
        \shortstack{\ding{51}\\(Noisy Sign Gradients)} & 
        \shortstack{\ding{51}\\(High)} & 
        High & 
        Low & 
        High \\
        \hline
        \shortstack{HE\\\cite{cheon2017}} & 
        \shortstack{Cryptographic\\(RLWE-based HE)} & 
        \shortstack{\ding{55}\\(Fully Encrypted)} & 
        \ding{55} & 
        \shortstack{Very\\Low} & 
        \shortstack{Very\\High} & 
        \shortstack{Very\\Limited} \\
        \hline
        \shortstack{\textsc{signSGD-MV}\\\cite{Ber18}} & 
        None & 
        \shortstack{\ding{51}\\(All Raw Sign Gradients)} & 
        \ding{55} & 
        \shortstack{Very\\High} & 
        \shortstack{Very\\Low} & 
        \shortstack{Very\\High} \\
        \hline
        \textbf{\shortstack{Proposed\\Method}} & 
        \textbf{\shortstack{Cryptographic\\(Beaver triples)}} & 
        \shortstack{\ding{51}\\\textbf{(Final Majority Vote Only})} & 
        \textbf{\ding{55}} & 
        \textbf{High} & 
        \textbf{Low} & 
        \textbf{High} \\
        \hline
    \end{tabular}
\end{table*}

% ==================================================================================

\section{Hi-SAFE Design} \label{sec:proposed method}

\subsection{Problem Setting and Design Criteria}
\label{sec:hisafe_design}

We design Hi-SAFE under the following FL environment. We adopt the semi-honest model, in which all users comply with the protocol, although some may attempt to infer private information from the exchanged messages~\cite{Bonawitz2017, zhang2023ahsecagg, zhao2023secretshared, jiang2024pqsf, liu2024mkfhe}. In addition, we employ the \textsc{signSGD-MV} update rule, whereby each user transmits only the 1-bit sign of its local gradient, and the server determines the global update direction by performing a majority vote over all received signs~\cite{Sei14,Ber18,Ber19,park2023,jin2024sign}. 

Based on this setting, Hi-SAFE is designed to achieve both communication efficiency and strong cryptographic privacy through the following core components.
\begin{enumerate}[leftmargin=1em]
    \item \textbf{Majority Vote Polynomial \(F(\xv)\) (see Section~\ref{subsec:mvpoly_scalar}):}  
    During the preprocessing stage, we construct a finite-field majority vote polynomial \(F(\xv)\) based on Fermat’s Little Theorem. This preprocessing step defines all coefficients of \(F(\xv)\) in advance, enabling the protocol to reproduce the standard majority vote result directly 
    during online execution, without evaluating or revealing any intermediate values.

    \item \textbf{Secure Polynomial Evaluation (see Section~\ref{Secure Evaluation MV_Beaver}):}  
    Each user securely evaluates its additive secret share of the polynomial \(F(\xv)\) without revealing its input \(\xv\). In this work, we adopt Beaver triples~\cite{beaver1991efficient} for secure multiplication, which mask the user's input and yield encrypted shares for aggregation; however, other secure multiplication techniques (e.g., DN~\cite{damgaard2007scalable} and ATLAS~~\cite{Goyal2021atlas}) can be seamlessly integrated into our framework.

    \item \textbf{Secure Aggregation and Broadcasting (see Section~\ref{Secure FL Framework}):}  
    The server aggregates the encrypted shares by summation to compute the final majority vote \(F(\xv)\), which is then broadcast to users for model update. Only the final result is revealed; all intermediate values remain hidden.

    \item \textbf{Hierarchical Aggregation via Subgrouping (see Section~\ref{sec:subgrouping}):}  
    To mitigate the overhead associated with secure polynomial evaluation using techniques such as Beaver triples, which grows significantly as the number of users increases, we divide users into subgroups that perform independent intra-subgroup aggregation. The final result is then obtained by aggregating the outputs of each subgroup, enabling both scalability and privacy while keeping the computational and communication costs manageable.
\end{enumerate}

The notations frequently used throughout this paper are summarized in Table~\ref{table:notation}. Standard mathematical symbols are employed unless otherwise specified.
\begin{table}[!htb]
\centering
\caption{Summary of notations}
\label{table:notation}
\renewcommand{\arraystretch}{1.2}
\begin{tabular}{|c|l|}
\hline
\textbf{Notation} & \textbf{Description} \\ \hline
\(\mathbb{F}_p\) & Finite field of prime order \(p\) \\ \hline
\(\mod p\) & Modulo operation over a prime \(p\) \\ \hline
\([n]\) & Index set \(\{1, 2, \dots, n\}\) \\ \hline
\(x, y\) & Scalars (denoted in regular lowercase letters) \\ \hline
\(\mathbf{x}, \mathbf{y}\) & Vectors (denoted in bold lowercase letters) \\ \hline
\(f(\thetav)\) & Global objective function evaluated at \(\thetav\) \\ \hline
\(f^\star\) & Minimum value of the global objective function \\ \hline
%\(\vec{a}\) & Non-negative vector \([a_1,\dots,a_d]\) \\ \hline
\(\vec{L}\) & Smoothness vector \([L_1,\dots,L_d]\) for function \(f\) \\ \hline
\(\vec{\sigma}\) & Variance bound vector for stochastic gradients \\ \hline
\(\llbracket x \rrbracket_i\) & Share of \(x\) for user \(i\) \\ \hline
\(\|\cdot\|_1\) & \(\ell_1\)-norm \\ \hline
\(C_u\) & Per-user communication cost \\ \hline
\(C_T\) & Total communication cost \\ \hline
\(R\) & Total number of secure multiplications \\ \hline
\(\mathbb{E}[\cdot]\) & Expectation operator for random variables \\ \hline
\end{tabular}
\end{table}

\subsection{Secure Evaluation of the Majority Vote Polynomial $F(\xv)$ over $\mathbb{F}_p^d$}
\label{subsec:secure_mv_polynomial}

This section describes the construction and secure evaluation of the majority vote polynomial \(F(\xv)\) over the $d$-dimensional vector space $\mathbb{F}_p^d$, where $\mathbb{F}_p$ denotes the prime field for a prime $p>n$. Each user provides a 1-bit sign gradient vector $\xv_i \in \{-1,+1\}^d$, and the primary objective is to compute the aggregated update direction (i.e., the majority votes) over these sign gradients while preserving individual user privacy under an honest-but-curious setting.

\subsubsection{Majority Vote Polynomial Construction via Fermat's Little Theorem}
\label{subsec:mvpoly_scalar}

Fermat’s Little Theorem allows the construction of an indicator polynomial over $\mathbb{F}_p$ that evaluates to $1$ when the input equals a target value and $0$ otherwise~\cite{ref4}.Let each user contribute a vector $\xv_i \in \{-1,+1\}^d$, and define the aggregate $\xv = \sum_{i=1}^{n} \xv_i \in \mathbb{F}_p^d$, where $p$ is the smallest prime greater than $n$. For any coordinate $j$, the scalar sum $x^{(j)} = \sum_{i=1}^n x_i^{(j)}$ lies in $\{-n,-n+2,\dots,n\}$. When \( n \) is even, a tie (\( \xv = 0 \)) may occur. Two common tie-breaking rules are:
\begin{itemize}[leftmargin=1.5em]
    \item \( \text{sign}(0) \in \{-1, +1\} \): tie resolved to binary decision (1-bit output of \( F(\xv) \)),
    \item \( \text{sign}(0) = 0 \): tie represented as a distinct third state (2-bit output of \( F(\xv) \)).
\end{itemize}

\vspace{1mm}
\noindent\textbf{Scalar Polynomial.}
Using Fermat’s Little Theorem, the term $1-(x^{(j)}-m)^{p-1}$ serves as an indicator for $x^{(j)}=m$. The majority vote for a scalar input $x$ is therefore represented exactly as
\begin{equation}
F(x)
=
\sum_{m\in\{-n,-n+2,\dots,n\}}
\text{sign}(m)\,
\big[1-(x-m)^{p-1}\big]
\pmod{p}.
\label{eq:mvpoly_scalar}
\end{equation}
where \( m =\sum_{i=1}^{n}m_i\) with \(m_i \in \{-1, +1\}\) and \( \text{sign}(0) \) is defined by the tie-breaking policy.

\vspace{1mm}
\noindent\textbf{Vector Extension.}
Since each coordinate is aggregated independently, the vector-valued polynomial is obtained component-wise:
\[
F(\xv)
=
\big(
F(x^{(1)}),
F(x^{(2)}),
\dots,
F(x^{(d)})
\big).
\]
\begin{comment}
\begin{lemma}[Correctness of the Majority Vote Polynomial]
Let $x=\sum_{i=1}^n x_i$ with $x_i\in\{-1,+1\}$ for all \( i \in [n] := \{1,2,...,n\} \). For any prime $p>n$, the polynomial $F(x)$ in~(\ref{eq:mvpoly_scalar}) satisfies $F(x) = \text{\emph{sign}}(x)$.
\end{lemma}

\begin{proof}
By Fermat’s Little Theorem, $(x-m)^{p-1} \equiv 0 \pmod{p}$ if $x=m$ and $1 \pmod{p}$ otherwise. Hence, in~(\ref{eq:mvpoly_scalar}), all terms vanish except the unique term with $x=m$, yielding $F(x)=\text{sign}(x)$. %Therefore, we obtain $F(\xv) \!=\! \text{sign}\left(\sum_{i=1}^n \xv_i\right) \!=\! \text{sign}(\xv)$, which coincides with the standard majority vote result.
\end{proof}
\end{comment}
% --------- 수정
\begin{lemma}[Correctness of the Majority Vote Polynomial]
Let $\xv = \sum_{i=1}^n \xv_i \in \mathbb{F}_p^d$, $\xv_i \in \{-1,+1\}^d$ for all \( i \in [n] \), and let $F(\xv) = \big(F(x^{(1)}),F(x^{(2)}),\dots,F(x^{(d)})\big)$ be the component-wise extension of the scalar polynomial \( F(x) \) defined in~(\ref{eq:mvpoly_scalar}). For any prime \( p > n \), the vector-valued polynomial satisfies
\[
F(\xv)=\mathrm{sign}(\xv),
\]
i.e., each coordinate of \( F(\xv) \) matches the standard majority vote result of \textsc{signSGD-MV}.
\end{lemma}

\begin{proof}
For any coordinate \( j\in\{1,\dots,d\} \), let $x^{(j)} = \sum_{i=1}^n x_i^{(j)}$, $x_i^{(j)} \in \{-1,+1\}$. Since \( x^{(j)} \in \{-n,-n+2,\dots,n\} \), exactly one value \( m \) in the summation of~(\ref{eq:mvpoly_scalar}) satisfies \( x^{(j)} = m \). By Fermat’s Little Theorem, 
\[
(x^{(j)} - m)^{p-1}
\equiv
\begin{cases}
0 \pmod{p}, & x^{(j)} = m, \\
1 \pmod{p}, & x^{(j)} \neq m.
\end{cases}
\]
Thus every term in~(\ref{eq:mvpoly_scalar}) vanishes except the unique term with \( x^{(j)}=m \), yielding $F(x^{(j)}) = \mathrm{sign}(x^{(j)})$.
Applying this argument independently to all coordinates,
\begin{align*}
    F(\xv) &=\big(\!F(x^{(1)}),\!\dots\!,F(x^{(d)}\!)\big)\\
    &=\big(\!\mathrm{sign}(x^{(1)}),\!\dots\!,\mathrm{sign}(x^{(d)})\big)\!=\!\mathrm{sign}(\xv),
\end{align*}
% \[
% F(\xv) \!=\! \big(\!F(x^{(1)}),\!\dots\!,F(x^{(d)}\!)\big)\!=\!\big(\!\mathrm{sign}(x^{(1)}),\!\dots\!,\mathrm{sign}(x^{(d)})\big)\!=\!\mathrm{sign}(\xv),
% \]
which coincides exactly with the majority vote result of \textsc{signSGD-MV}.
\end{proof}

%} % end of RED
%\end{comment}

%%%%%%%%%%%%%%%%%%%%%%%%%%%%%%%%%%%%%%%%%%%%%%%%%%%%%%%%%%%%%%%%%%%%%%%%%%%%

Once the number of users $n$ and the tie-breaking policy are specified, the majority vote polynomial $F(\xv)$ can be systematically constructed and efficiently precomputed according to Eq.~(\ref{eq:mvpoly_scalar}) during the offline phase. 
Table~\ref{table:majority_vote} presents representative examples of precomputed polynomials according to tie-breaking policies. 

\begin{table}[!htb]
    \centering
    %\small
    \setlength{\tabcolsep}{2pt}
    \caption{Precomputed majority vote polynomials \( F(\xv) \) according to tie-breaking policies}
    \label{table:majority_vote}
    \begin{tabular}{|c|c|c|}
    \hline
    \#Users & \( \text{sign}(0)\in \{-1, +1\} \) & \( \text{sign}(0) = 0 \) \\ \hline 
    $n=2$ & \( \xv^2 + 2\xv + 2 \pmod{3} \) & \(2\xv \pmod{3} \) \\ 
    $n=3$ & \( 2\xv^3 + 4\xv \pmod{5} \) & \( 2\xv^3 + 4\xv \pmod{5} \) \\
    $n=4$ & \( \xv^4 + 3\xv^3 + \xv + 4 \pmod{5} \) & \( 3\xv^3 + \xv \pmod{5} \) \\
    $n=5$ & \( 3\xv^5 + 2\xv^3 + 3\xv \pmod{7} \) & \( 3\xv^5 + 2\xv^3 + 3\xv \pmod{7} \) \\
    $n=6$ & \( \xv^6 + 4\xv^5 + 5\xv^3 + 4\xv + 6 \pmod{7} \) & \( 4\xv^5 + 5\xv^3 + 4\xv \pmod{7} \) \\ \hline
    \end{tabular}
\end{table}

%To perform secure multiplications during polynomial evaluation, we employ additive secret sharing techniques, for example using Beaver triples, as a practical realization. An offline phase allows users to jointly prepare the required randomness without server involvement (DN, ATLAS)~\cite{beaver1991efficient, damgaard2013practical, rathee2020cryptflow2}.
%\subsubsection{Secure Evaluation of Majority Vote Polynomial $F(\xv)$ Using Beaver Triples}\label{Secure Evaluation MV_Beaver}

% ======================================= 원본
\subsubsection{Secure Evaluation of Majority Vote Polynomial $F(\xv)$}\label{Secure Evaluation MV_Beaver}

In the FL setting, each user holds a private input \(\xv_i\) (e.g., sign gradient), and the goal is to securely evaluate a majority vote polynomial \(F(\xv)\) over the aggregated value \(\xv \!=\! \sum_{i=1}^{n} \xv_i\), without revealing any input \(\xv_i\). 
%To perform secure multiplications during polynomial evaluation, we utilize Beaver triples under an additive secret sharing scheme. 
To perform secure multiplications during polynomial evaluation, we employ additive secret sharing techniques, instantiated for example via Beaver triples~\cite{beaver1991efficient}, as a practical realization. 

For simplicity, we omit the \!\!\(\pmod{p}\) operation and the explicit coefficients of \(F(\xv)\). Let $\deg(F)$ denote the degree of \(F(\xv)\) over $\mathbb{F}_p$. In the offline (initialization) phase, the users collaboratively generate Beaver triples $\{(\llbracket \av^r \rrbracket_i, \llbracket \bv^r \rrbracket_i, \llbracket \cv^r \rrbracket_i): r \in [R]\}$ via MPC, and each user locally retains its own share, where $R$ is the number of multiplications for securely evaluating the majority vote polynomial. 
% ============== x^k 정의 추가
During the online phase (\textit{subround}) for secure polynomial evaluation, each user $i$ recursively computes the shares $\llbracket \xv^k \rrbracket_i$ of powers $\xv^k$ for $k\!=\!1,2,\!\dots,\deg(F)$ as
\begin{equation}
\llbracket \xv^{k} \rrbracket_i \!=\!
\begin{cases}
   \xv_i, &\!\! k \!=\! 1, \\[5pt]
   \llbracket \cv^r \rrbracket_i 
   \!+\! \deltav_{k-v_k}^r\! \cdot \! \llbracket \bv^r \rrbracket_i 
   \!+\! \epsilonv_{v_k}^r\! \cdot \! \llbracket \av^r \rrbracket_i 
   \!+\! \deltav_{k-v_k}^r\! \cdot \! \epsilonv_{v_k}^r, &\!\! k \!>\! 1,
\end{cases}
\label{eq:share of power}
\end{equation}
where $v_k=\max\{j\in\mathbb{N}\mid 2^j \leq k-1\}$ and $(\deltav_{k-v_k}^r,\epsilonv_{v_k}^r)$ are obtained by aggregating the masked differences. A fresh Beaver triple is consumed for each multiplication, ensuring that higher-order terms of $F(\xv)$ are securely computed without exposing any individual input.
% ===================================================================

The \textit{\textbf{subround procedure}} for the secure evaluation of $F(\xv)$ within the FL framework is as follows:

%\noindent\textbf{Step 1) (Offline Phase) Beaver Triple Generation:}  
%Server~2 generates and distributes Beaver triples \(\{(\llbracket \av^r \rrbracket_i, \llbracket \bv^r \rrbracket_i, \llbracket \cv^r \rrbracket_i): r \in [R]\}\) to each user $i$, where $\llbracket \av^r \rrbracket_i, \llbracket \bv^r \rrbracket_i, \llbracket \cv^r \rrbracket_i \in \mathbb{F}_p, \cv^r=\av^r\cdot\bv^r$ and \( R \) denotes the number of multiplications required to securely evaluate the majority vote polynomial.

\noindent\textbf{Step 1) Evaluation of Shares $\llbracket \xv^k \rrbracket_i$ for \(k = 2\) to \(\deg(F)\):}
\begin{itemize}[leftmargin=0.0em]
    \item For each \(k\), each user \(i\) computes the masked differences \(\llbracket \xv^{k-v_k} \rrbracket_i - \llbracket \av^r \rrbracket_i\) and \(\llbracket \xv^{v_k} \rrbracket_i - \llbracket \bv^r \rrbracket_i\) based on Eq.~(\ref{eq:share of power}), and sends them to Server.
 
    \item Server aggregates the received masked values to compute:
    $\deltav_{k-v_k}^r \!\!=\!\! \sum_{i=1}^{n} (\llbracket \xv^{k-v_k} \rrbracket_i \!-\! \llbracket \av^r \rrbracket_i) \!=\! \xv^{k-v_k} \!-\! \av^r$ \text{ and }
    $\epsilonv_{v_k}^r \!\!=\!\! \sum_{i=1}^{n} (\llbracket \xv^{v_k} \rrbracket_i \!-\! \llbracket \bv^r \rrbracket_i) \!=\! \xv^{v_k} \!-\! \bv^r$, and broadcasts both \!\(\deltav_{k-v_k}^r\)\! and \!\(\epsilonv_{v_k}^r\) \!to all users.\!\!\!
\end{itemize}
%\vspace{1mm}
\noindent\textbf{Step 2) Local Polynomial Encryption:} Using all received pairs\! \(\{(\deltav_{k-v_k}^r, \epsilonv_{v_k}^r)\!:\!k=2, \!..., \!\deg(\!F), r\! \in \![R]\}\), each user \(i\) locally computes its encrypted share of the evaluated $F(\xv)$ as:
%-------------- 수정
\begin{align}
\label{eq_share}
Enc(\xv_i) 
&= \llbracket F(\xv) \rrbracket_i 
= \sum_{k=2}^{\deg(F)} \sum_{r=1}^{R} \Big(
    \llbracket \cv^r \rrbracket_i
    + \deltav_{k-v_k}^r \cdot \llbracket \bv^r \rrbracket_i \nonumber \\ 
&\qquad
    +\, \epsilonv_{v_k}^r \cdot \llbracket \av^r \rrbracket_i + \deltav_{k-v_k}^r \cdot \epsilonv_{v_k}^r
\Big) + \xv_i \pmod{p}.
\end{align}
\begin{comment}
\noindent\textbf{Step 3) (Online Phase - Global Round) Secure Evaluation of $F(\xv)$ on Server:} 
Server aggregates all encrypted shares to compute the final majority vote result without revealing any user’s input \(\xv_i\): %The aggregation proceeds as:
\begin{align}
F(\xv)\!=\!\! \sum_{i=1}^{n} Enc(\xv_i) \!=\!\! \sum_{i=1}^{n} \llbracket F(\xv) \rrbracket_i \pmod{p}. 
\label{eq_share_sum}
\end{align}
\end{comment}
The overall encryption procedure is summarized in Algorithm~\ref{alg_enc}, which covers only the user-side encryption steps based on Beaver triples according to the subround in FL framework. %The final aggregation performed by the server is handled separately as a simple summation of encrypted shares. 
For a concrete illustration, see Appendix~\ref{sec:toy example}.
\begin{algorithm}[!tb]
\caption{Encryption of Majority Vote Polynomial $F(\xv)$ via Secure Multiplication (\textit{Subround})}
\begin{algorithmic}[1]
\STATE \textbf{Input:}  \# selected users $n$, majority vote polynomial $F(\xv)$, \# multiplications $R$

%\STATE \textbf{Offline Phase [on Server~2]}: Initialization
%\STATE \hspace{0.2cm} \textbf{construct} majority vote polynomial $F(\xv)$ using Eq.~(\ref{eq2}).
%\STATE \hspace{0.2cm} \textbf{generate} and \textbf{distribute} Beaver triples \(\{(\llbracket \av^r \rrbracket_i, \llbracket \bv^r \rrbracket_i, \llbracket \cv^r \rrbracket_i): r \in [R]\}\) \textbf{to} each user $i$.

\STATE \textbf{Online Phase}: Encryption of majority vote polynomial $F(\xv)$ 
\STATE \hspace{0.2cm} \textbf{for} $k = 2$ to $\deg(F)$ \textbf{do} \hfill (subrounds for evaluating the shares)
\STATE \hspace{0.5cm} \textbf{[On User $i$]} \textbf{compute} $\llbracket \xv^{k-v} \rrbracket_i - \llbracket \av^r \rrbracket_i$ and $\llbracket \xv^v \rrbracket_i - \llbracket \bv^r \rrbracket_i$, and \textbf{send} them \textbf{to} Server.
\STATE \hspace{0.5cm} \textbf{[On Server]} \textbf{aggregate} masked values to obtain $\deltav_k^r$, $\epsilonv_k^r$, and \textbf{broadcast} them \textbf{to} all users.
\STATE \hspace{0.2cm} \textbf{end for}
\STATE \hspace{0.2cm} \textbf{[On User $i$]} \textbf{compute} secret share $\llbracket F(\xv) \rrbracket_i$ using Eq.~(\ref{eq_share}).
\STATE \textbf{Output:} (Generation of $\llbracket F(\xv) \rrbracket_i$ for user $i$) $Enc(\xv_i) = \llbracket F(\xv) \rrbracket_i$
\end{algorithmic}
\label{alg_enc}
\end{algorithm}

% =========================================================================================

%\vspace{-1mm}
\subsection{Secure Multiplication-Based FL Framework} \label{Secure FL Framework}
%\vspace{-1mm}
In this section, we introduce a novel FL framework that integrates secure aggregation via secure multiplications to preserve user privacy while maintaining aggregation correctness.
To clarify the internal mechanisms of the proposed framework, we first describe the key update procedures executed by the users and the central server, respectively.
%\vspace{-1mm}
\subsubsection{User Update Procedure} 
%\vspace{-1mm}
\textbf{Step 1 (Sign Gradient Calculation): } 
The user $i$ computes the gradient $\gv_i(t)$ using the global model $\thetav(t)$ and performs 1-bit quantization to obtain the locally updated sign gradient $\xv_{i}(t)$:
\begin{equation}
\xv_{i}(t) = \text{sign}(\gv_i(t)), \quad \xv_{i}(t) \in \{-1,1\}^d
\end{equation}
where $d$ denotes the size of the global model.

\textbf{Step 2 (Secure Evaluation of $\llbracket F(\xv) \rrbracket_i$):}  
%Each user~$i$ employs Beaver triples that are pre-distributed by Server~2 to securely evaluate a share of the majority vote polynomial \( F(\xv(t)) \) for $\xv(t)\!=\!\!\sum^n_{i=1} \xv_i(t)$, represented as $\llbracket F(\xv(t)) \rrbracket_i$. This share is utilized to compute the sign of the aggregated input vectors.
At subround, each user~$i$ employs Beaver triples, as an example instantiation of secure multiplication, pre-distributed to securely evaluate its share of the majority vote polynomial \( F(\xv(t)) \) for $\xv(t)\!=\!\!\sum^n_{i=1} \xv_i(t)$, represented as $\llbracket F(\xv(t)) \rrbracket_i$. This share is then used to compute the sign of the aggregated input vectors. 
%Further details are provided in Section~\ref{Secure Computation of MV polynomials} and Algorithm~\ref{alg_enc}. 
Notably, this computation is performed without revealing any individual input $\xv_i(t)$.  
Further details are provided in Section~\ref{subsec:secure_mv_polynomial} and Algorithm~\ref{alg_enc}. Finally, each user $i$ sends its encrypted share of the majority vote polynomial to the server: $Enc(\xv_i(t)) = \llbracket F(\xv(t)) \rrbracket_i$.
%%%%%%%%%%%%%%%%%%%%%%%%%%%%%%%%%%%%%%%%%%%%%%%%%%%%%%%%%%%%%
%\vspace{-1mm}
\subsubsection{Model Aggregation Procedure}
%\vspace{-1mm}
From the encrypted local updates $\{Enc(\xv_i(t)): i \in [n]\}$, Server computes the final majority vote result \( \tilde{\gv}(t) \) and broadcasts it to all users as follows:
%\vspace{-1mm}

\textbf{Aggregation: }
\begin{align} 
      F(\xv(t))\!=\!\!\!\sum_{i=1}^n Enc(\xv_i(t)) \!=\!\!\! \sum_{i=1}^n \llbracket F(\xv(t)) \rrbracket_i\!\!\!\!\! \pmod{p}, %\hspace{2em}
\label{eq_agg}
\end{align} 
where $\xv(t)\!=\!\!\sum^n_{i=1} \xv_i(t)$ and $\sum_{i=1}^n \llbracket F(\xv(t)) \rrbracket_i \!=\!\text{sign}(\xv(t))$. 
%\vspace{1mm}
\begin{align}
\textbf{Broadcasting:  }\quad\quad\quad
     \tilde{\gv}(t) = F(\xv(t)) = \text{sign}(\xv(t)). %\hspace{7.5em}
\end{align}

The overall aggregation procedure is summarized in Algorithm~\ref{alg_SA}.
Each user encrypts and transmits its share of the majority vote polynomial \( F(\xv) \), while the server aggregates the received shares as in Eq.~(\ref{eq_agg}) and broadcasts the resulting global direction \( \tilde{\gv}(t) \) to all users for model update.

%%%%%%%%%%%%%%%%%%%%%%%%%% Secure aggregation using Beaver triple
\begin{algorithm}[!htb]
\caption{Secure Majority Vote Aggregation via Secure Multiplication}
\label{alg_SA}
\begin{algorithmic}[1]
\STATE {\textbf{Input:}}  Initial model $\thetav_0$, learning rate $\eta$, \# selected users $n$, majority vote polynomial $F(\xv)$
\FOR{$t = 0$ to $T-1$}
    \STATE \textbf{[On User $i$]}
    \STATE \hspace{0.5em} \textbf{compute} local gradient: $\gv_{i}(t)$
    \STATE \hspace{0.5em} \textbf{quantize} gradient: $\xv_{i}(t) = q(\gv_{i}(t)) \in \{-1,1\}^d$
    \STATE \hspace{0.5em} \textbf{generate} secret share: $Enc(\xv_{i}(t)) \!\gets\! \llbracket F(\xv(t)) \rrbracket_i$  using Algorithm~\ref{alg_enc} for $\xv(t)\!=\!\!\sum^n_{i=1} \xv_i(t)$ 
    \STATE \hspace{0.5em} \textbf{transmit} $Enc(\xv_{i}(t))$ \textbf{to} Server
    \STATE \textbf{[On Server]}
%%%%%%%%%%%%%%%
    \STATE \hspace{0.5em} \textbf{aggregate} encrypted shares:$F(\xv(t)) = \sum_{i=1}^n Enc(\xv_i(t))$ \hfill (see Eq.~(\ref{eq_agg}))
    \STATE \hspace{1em} \textbf{obtain} majority vote: $\tilde{\gv}(t) = \text{sign} \left( \sum_{i=1}^n \xv_i(t) \right) \gets F(\xv(t))$
%%%%%%%%%%%%%%
    \STATE  \hspace{0.5em} \textbf{broadcast} $\tilde{\gv}(t)$ \textbf{to} all users
    \STATE \textbf{[On User $i$]} \textbf{update} model: $\thetav(t+1) \gets \thetav(t) - \eta \tilde{\gv}(t)$
\ENDFOR
\STATE {\textbf{Output:}} $\thetav(T)$
\end{algorithmic}
\end{algorithm}

%%%%%%%%%%%%%%%%%%%%%%%%%%%%%%%%%%%%%%%%%%%%%%%%%%%%%%%%%%%%%%%%%%%%%%%%%%%%%%%%%%
%%%%%%%%%%% subgrouping
%%%%%%%%%%%%%%%%%%%%%%%%%%%%%%%%%%%%%%%%%%%%%%%%%%%%%%%%%%%%%%%%%%%%%%%%%%%%%%
\subsection{Subgroup-Based Secure FL Framework} \label{sec:subgrouping}
 
To mitigate the overhead of securely evaluating the majority vote polynomial \(F(\xv)\) with secure multiplication techniques (e.g., Beaver triples, DN, ATLAS), whose cost grows significantly with the number of users, we propose a subgrouping strategy that partitions users into smaller subsets \(\mathcal{G}_j\). Each subgroup securely aggregates its inputs independently, and the final result is obtained by combining all subgroup outputs. This reduces the polynomial degree, latency, and bandwidth, while ensuring scalability and privacy with manageable computational and communication costs.

\noindent\textbf{Subgrouping and Hierarchical Majority Vote Aggregation: }
As the number of users \( n \) increases, the degree of the majority vote polynomial \( F(\xv) \) also grows, which raises the number of secure multiplication subrounds required for polynomial evaluation. This results in higher uplink communication cost and latency, thereby limiting scalability. 
In addition, a larger prime modulus \(p\!>\!n\) must be chosen, which further increases the complexity of evaluating \(F(\xv)\) over \(\mathbb{F}_p\).
To address these limitations, we propose a subgrouping strategy that partitions the total \( n \) users into \( \ell \) disjoint subgroups, each of size \( n_1 \!=\! n / \ell \). Within each subgroup, a small majority vote polynomial is evaluated independently based on local inputs. Since the polynomial degree now depends on the smaller subgroup size \( n_1 \), the number of required secure subrounds is reduced and a smaller prime modulus \( p_1 (>n_1) \) is adopted. Consequently, subgrouping leads to significant reductions in both computational and communication costs. %{\RED However, subgrouping exposes each subgroup’s majority vote sign to the server, limiting privacy to the subgroup level. By adding protection to the subgroup outputs, we can restore full end-to-end privacy so that only the global majority vote result is revealed.} %We further analyze the impact of tie-breaking policies and hierarchical aggregation on communication and computation, as detailed in Appendix~\ref{appendix:tie-breaking policy}.}

The proposed aggregation procedure is executed in two hierarchical steps:

\noindent\textbf{Step 1 (Intra-subgroup Majority Vote):}  
Within each subgroup \(\mathcal{G}_j, j \in [\ell]\), %the server securely evaluates the local majority vote as
the local majority vote is securely evaluated as
    \begin{equation}
    \label{eq:enc_sum}
    F(\xv_j(t))=\sum_{i=1}^{n_1} Enc(\xv_{i,j}(t)) = \sum_{i=1}^{n_1} \llbracket F(\xv_{j}(t)) \rrbracket_i \pmod{p_1},
    \end{equation}
    where $\sum_{i=1}^{n_1} \llbracket F(\xv_{j}(t)) \rrbracket_i \!=\! \text{sign} \left( \xv_{j}(t) \right)$ for $\xv_{j}(t)\!\!=\!\!\sum_{i=1}^{n_1} \xv_{i,j}(t)$.

\noindent\textbf{Step 2 (Inter-subgroup Majority Vote):}  
The global majority vote is computed by aggregating the results across all subgroups:
    \begin{equation}
    \label{eq:inter_mv}
    \tilde{\gv}(t) = \text{sign}\bigg( \sum_{j=1}^{\ell} F\big(\xv_j(t)\big) \bigg) = \text{sign} \bigg( \sum_{j=1}^{\ell} \text{sign} \bigg( \sum_{i=1}^{n_1} \xv_{i,j}(t) \bigg) \bigg).
    \end{equation}
%\noindent\textbf{Broadcasting:}  
The global majority vote result \(\tilde{\gv}(t)\) is subsequently broadcast to all users.
% =========================================================================
The overall protocol is summarized in Algorithm~\ref{alg:secure_mv}.
\begin{algorithm}[!htb]
\caption{Hierarchical Secure Majority Vote Aggregation with Subgrouping}
\label{alg:secure_mv}
\begin{algorithmic}[1]
\STATE {\textbf{Input:}} Initial model $\thetav_0$, learning rate $\eta$, \# selected users $n$, \# subgroups $\ell$, majority vote polynomial $F(\xv_j)$ for each subgroup 
\FOR{$t = 0$ to $T-1$}
    \STATE \textbf{[On User $i$ in subgroup $\mathcal{G}_j$]}
    \STATE \hspace{0.5mm} \textbf{compute} local gradient: $\gv_{i,j}(t)$
    \STATE \hspace{0.5mm} \textbf{quantize} gradient: $\xv_{i,j}(t) = q(\gv_{i,j}(t))\in \{-1,1\}^d$
    \STATE \hspace{0.5mm} \textbf{generate} secret share: \!$Enc(\xv_{i,j}(t)\!) \!\gets\! \llbracket \!F(\xv_j(t))\! \rrbracket_i$ for $\xv_{j}(t)\!=\!\!\sum_{i=1}^{n_1}\! \xv_{i,j}(t)$ using Algorithm~\ref{alg_enc}\!\!\!\!
    \STATE \hspace{0.5mm} \textbf{transmit} $Enc(\xv_{i,j}(t))$ \textbf{to} Server
    \STATE \textbf{[On Server]}
    \STATE \hspace{0.5mm} \textbf{reconstruct} $F(\xv_j(t))$ \textbf{from} received shares for each subgroup $\mathcal{G}_j$ \hfill (see Eq.~(\ref{eq:enc_sum}))
    \STATE \hspace{0.5mm} \textbf{compute} global vote: $\tilde{\gv}(t) = \text{sign} ( \sum_{j=1}^{\ell} F(\xv_j(t)) )\in \{-1,1\}^d$
    \STATE \hspace{0.5mm} \textbf{broadcast} $\tilde{\gv}(t)$ \textbf{to} all users
    \STATE \textbf{[On User $i$]} \textbf{update} model: $\thetav(t+1) \gets \thetav(t) - \eta \tilde{\gv}(t)$
\ENDFOR
\STATE {\textbf{Output:}} $\thetav(T)$
\end{algorithmic}
\end{algorithm}

\subsection{Tie-Breaking Policies for Hierarchical Majority Voting}
\label{sec:tie-breaking policy}
In the proposed subgroup-based FL framework, the choice of tie-breaking rule affects both the computational resolution of the majority vote polynomial and the communication cost. Two levels of majority voting are involved—\emph{intra-subgroup} and \emph{inter-subgroup}—and each may adopt different tie-handling schemes.

\textbf{Intra-Subgroup Majority Vote:}
\begin{itemize}[leftmargin=2em]
    \item \textbf{Case A:} $\text{sign}(0) \in \{-1,1\}$ (1-bit representation)
    \item \textbf{Case B:} $\text{sign}(0)=0$ (3-state output, requiring 2 bits)
\end{itemize}
Case~B increases the local computational resolution but does not affect uplink communication, since intra-subgroup computations remain internal to the server.

\textbf{Inter-Subgroup Majority Vote:}
\begin{itemize}[leftmargin=2em]
    \item \textbf{Case 1:} $\text{sign}(0) \in \{-1,1\}$ (1-bit downlink)
    \item \textbf{Case 2:} $\text{sign}(0)=0$ (2-bit downlink)
\end{itemize}

\textbf{Combined Configurations:}
\begin{itemize}[leftmargin=2em]
    \item \textbf{A–1 (1-bit tie-breaking):} 1-bit/1-bit (minimal communication overhead)
    \item \textbf{B-1 (2-bit tie-breaking):} 2-bit/1-bit (higher local resolution without additional communication
    \item A–2: 1-bit/2-bit (larger downlink)
    \item B–2: 2-bit/2-bit (maximum resolution and cost)
\end{itemize}

\textbf{Remark:}  
Configurations using Case~2 are incompatible with \textsc{signSGD-MV}, which assumes a 1-bit global update. Hence, we focuses on Cases A–1 and B–1, which satisfy the 1-bit constraint.
% -------------------------------------------------

% ===============================================================================
\section{Theoretical Analysis}

In this section, we present an analysis of the convergence, the security properties, and the computation and runtime overhead of the proposed Hi-SAFE method. The main theoretical results are proved in Appendices~\ref{appendix:hierarchical_proof} and~\ref{appendix:security_proofs}.%, which also provide a formal proof of corruption tolerance along with additional proofs that support the main analysis. %{\RED Furthermore, we provide a computational complexity analysis and runtime evaluation of the proposed method in Appendix~\ref{subsec:sv_complexity}.}

\subsection{Convergence Analysis}

We analyze the convergence of Hi-SAFE, a \textsc{SignSGD} algorithm with hierarchical majority vote. The analysis follows the structure of \cite{Ber18} and extends the original majority-vote convergence guarantee to the case where $n$ users are partitioned into $\ell$ subgroups of size $n_1 = n/\ell$.

Let $f^\star$ denote the minimum of the global objective, let $f_0$ be the initial objective value, and let $\vec{L}$ and $\vec{\sigma}$ denote the
coordinate-wise smoothness and variance vectors, respectively. Suppose each subgroup produces a coordinate-wise majority vote that is correct with
probability strictly greater than $1/2$, independently across subgroups.

\begin{theorem}[Convergence of \textsc{SignSGD} with Hierarchical Majority Vote]
\label{thm:hier_sigsgd}
Run Algorithm~\ref{alg:secure_mv} for $K$ iterations with step size $\eta = 1/\sqrt{K \|\vec{L}\|_1}$ and mini-batch size $m_k = K$, and let
$N_t = K^2$ denote the total number of stochastic gradient evaluations per user. Then the averaged gradient norm satisfies
\[
\begin{aligned}
\mathbb{E}\!\left[\frac{1}{K}\sum_{k=0}^{K-1}\|\gv_k\|_1\right]^2
&\!\!\le \! \frac{1}{\sqrt{N_t}}
\Bigg(\!
\sqrt{\|\vec{L}\|_1}(f_0 \!-\! f^\star \!+\! \tfrac12)
\!+\! \frac{2}{\sqrt{n_1}}\|\vec{\sigma}\|_1 \\
&\qquad\qquad\qquad
+\, C_{\mathrm{hier}} e^{-c_2 \ell}
\Bigg)^2.
\end{aligned}
\]
where $C_{\mathrm{hier}} = \sum_{j=1}^d \mathbb{E}[\,|g_{k,j}|\,]$, $c_2 = \frac{(2q - 1)^2}{2}$, and $q > \tfrac12$ denotes the minimum per-subgroup success probability (i.e., the probability that each subgroup majority vote matches the true sign).
\end{theorem}

\begin{remark}[Convergence–Communication Trade-off]
Larger subgroups (larger $n_1$ and smaller $\ell$) achieve lower variance and faster convergence. Conversely, increasing $\ell$ reduces per-user communication and enables scalable deployment. The exponentially suppressed global error guarantees that the loss in convergence due to subgrouping remains negligible for moderate~$\ell$.
\end{remark}

% ------- subgrouping에서 s_j leakage 경우
\subsection{Security Analysis}
\label{sec:security_analysis}

We analyze the privacy of Hi-SAFE under the semi-honest adversarial model. The goal of the protocol is to protect all individual user inputs and all intermediate arithmetic values, while revealing only \emph{aggregate sign information} in the form of subgroup majority votes and the final global majority.

More precisely, the $n$ users are partitioned into $\ell$ subgroups $\{\mathcal{G}_j\}_{j=1}^{\ell}$, each of size $n_1 = n/\ell$. User $i\in\mathcal{G}_j$ holds an input vector $\xv_{i,j}\in\{-1,+1\}^d$, and the subgroup computes $\xv_j = \sum_{i=1}^{n_1} \xv_{i,j}$, $\sv_j = \mathrm{sign}(\xv_j) \in \{-1,0,+1\}^d$. At the second layer, the server computes the global majority $\sv = \mathrm{sign}\Big(\sum_{j=1}^{\ell} \sv_j\Big) \in \{-1,0,+1\}^d$. Both the subgroup majority function $F(\xv_j)$ and the inter-group aggregation are evaluated via finite-field arithmetic operations, using Beaver triples whose masks are generated in an offline MPC phase and are independent of all inputs. The adversary corrupts at most $t\le n-1$ users and is semi-honest, i.e., follows the protocol but tries to infer additional information from its view.

At the intra-subgroup stage, the majority vote polynomials $\{F(\xv_j)\}$ are securely evaluated under additive secret sharing; the server then reconstructs the subgroup outputs $\{\sv_j\}_{j=1}^{\ell}$. At the inter-group stage, the server computes the final majority $\sv = \mathrm{sign}(\sum_{j=1}^{\ell} \sv_j)$ and broadcasts $\sv$ to all users.

We denote by $\mathsf{REAL}_{\mathcal{A}}$ the adversary’s view, including its corrupted inputs, randomness, all protocol messages, and the reconstructed subgroup and global majority votes.  By contrast, $\mathsf{SIM}_{\mathcal{A}}$ denotes the output of a PPT simulator that has access only to the corrupted inputs and to the \emph{leakage} $\{\sv_1,\dots,\sv_\ell,\;\sv\}$.

\begin{theorem}[Security of Hi-SAFE with Subgroup Majority Leakage]
\label{thm:hierarchical_secure}
Let $\mathcal{A}$ be any PPT semi-honest adversary corrupting a set
$\mathcal{C}\subseteq[n]$ of at most $t\le n-1$ users.  Let
$\{\sv_j\}_{j=1}^{\ell}$ denote the subgroup majority votes and let
$\sv$ denote the final global majority.  Then there exists a PPT
simulator $\mathsf{SIM}$ such that
\[
\mathsf{REAL}_{\mathcal{A}}\big(\{\xv_{i,j}\}_{i\in\mathcal{C}}\big)
\;\approx_c\;
\mathsf{SIM}_{\mathcal{A}}\big(\{\xv_{i,j}\}_{i\in\mathcal{C}},\,
\{\sv_j\}_{j=1}^{\ell},\,\sv\big),
\]
where $\approx_c$ denotes computational indistinguishability.
In particular, Hi-SAFE leaks no information beyond the corrupted user
inputs and the aggregate sign information 
$\{\sv_j\}_{j=1}^{\ell}$ and $\sv$.
\end{theorem}

\begin{remark}[Granularity of Leaked Information]
Theorem~\ref{thm:hierarchical_secure} formalizes privacy relative to the leakage $\mathsf{leak} = \{\sv_1,\dots,\sv_\ell,\sv\}$.  Intuitively, the server learns, for each coordinate, the majority sign within each subgroup and the majority sign across all subgroups, but no additional information about any individual user’s sign gradient beyond what is logically implied by these aggregate signs.
\end{remark}

\begin{remark}[Comparison with Flat Majority Vote]
In the flat (non-subgrouped) setting, the leakage reduces to a single vector $\sv = \mathrm{sign}(\sum_{i=1}^n \xv_i)$, matching the standard \textsc{SignSGD-MV} scenario. With subgrouping, Hi-SAFE additionally reveals the intermediate subgroup majorities $\{\sv_j\}$, which provide a finer-grained view of local aggregates but still do not expose any raw user inputs or intermediate arithmetic values of the MPC execution.
\end{remark}

% ----------------- 추가
\begin{remark}[Residual Leakage Probability]
For a single coordinate, the only case in which the final majority vote reveals all $n$ inputs is when all inputs are identical, which occurs with probability $2^{-(n-1)}$ in the flat setting and $2^{-(n_1-1)}$ under subgrouping. For a $d$-dimensional model, the corresponding probabilities become $(2^{-(n-1)})^{d}$ and $(2^{-(n_1-1)})^{d}$, respectively, both of which are negligible in practice.
\end{remark}

\subsubsection*{Discussion on Residual Leakage Probability}
If each input $\xv_{i,j}$ is independently and uniformly drawn from $\{-1,+1\}$, then observing the final majority sign $\sv$ fully reveals all inputs only when the input vectors are identical. In the flat (non-subgrouping) case, this event occurs with probability
\[
\Pr[\text{all inputs identical}] = 2^{-(n-1)}.
\]
Extending this over $d$ coordinates yields the model-level leakage probability
\[
\Pr[\text{privacy failure}] = (2^{-(n-1)})^{d},
\]
which decays doubly exponentially in $n$ and $d$.

When subgrouping is applied, each coordinate leakage depends only on the subgroup size $n_1$, producing
\[
(2^{-(n_1-1)})^{d}.
\]
Although $2^{-(n_1-1)} > 2^{-(n-1)}$ for $n_1<n$, the model-level leakage remains negligible because $(2^{-(n_1-1)})^{d}\ll 1$ for any practical dimension. Thus, subgrouping introduces a tunable privacy–efficiency trade-off: smaller $n_1$ reduces communication and secure-evaluation cost, whereas larger $n_1$ further suppresses the already negligible residual-leakage probability.

\subsection{Analysis of Computation and Runtime Overhead}
\label{subsec:sv_complexity}

\subsubsection{Computational Complexity of the Majority Vote Polynomial}
\label{subsec:mv-complexity}

We analyze the computational cost of constructing the majority vote polynomial \(F(\xv)\), whose form is given in Eq.~(\ref{eq:mvpoly_scalar}). A direct construction requires modular exponentiations of the form \((\xv-\mv)^{p-1}\), each computable in \(\mathcal{O}(\log p)\), yielding the overall naive complexity $\mathcal{O}(n\log p)$.

\paragraph*{Reduction via Subgrouping}
Under subgrouping, users are partitioned into \(\ell\) groups of size \(n_1=n/\ell\), and each group evaluates a smaller polynomial over \(\mathbb{F}_{p_1}\) with \(p_1>n_1\). The resulting cost becomes $\mathcal{O}(n_1\log p_1)$, and the polynomial is precomputed once offline and reused across global rounds.

\begin{table}[h]
\centering
\caption{Computational complexity of majority vote polynomial construction}
\label{tab:complexity}
\begin{tabular}{|c|c|}
\hline
\textbf{Method} & \textbf{Total Complexity} \\
\hline
Non-subgrouping & $\mathcal{O}(n\log p)$ \\
With subgrouping & $\mathcal{O}(n_1\log p_1)$ \\
\hline
\end{tabular}
\end{table}

These reductions make the computation lightweight and scalable, enabling efficient secure aggregation in large-scale FL settings.

\subsubsection{Computation and Runtime Overhead in Secure Evaluation}
\label{subsec:sv_complexity}

To assess the efficiency of the proposed secure evaluation scheme (Algorithm~1), we measured both the offline preprocessing cost (Beaver triple generation) and the online execution cost for secure polynomial evaluation under the adopted subgroup configuration. The results are summarized in Table~\ref{tab:complexity_time}.
\begin{table}[!htb]
\centering
\setlength{\tabcolsep}{3pt}
\caption{Runtime and computational complexity of Algorithm~1.}
\label{tab:complexity_time}
\begin{tabular}{lccc}
\toprule
\textbf{Phase} & \textbf{Operation} & \textbf{Complexity} & \textbf{Average Runtime (s)} \\
\midrule
Offline & Beaver triple generation
& $\Theta\!\big(\ell\, d_{\mathrm{sub}}\, n_1^2\big)$
& $< 0.01$ \\
Offline & Precomputation of $F_{\mathrm{sub}}$
& $\mathcal{O}(n_1 \log p_1)$
& $< 0.01$ \\
Online  & Secure evaluation of $F_{\mathrm{sub}}$
& $\Theta\!\big(\ell\, d_{\mathrm{sub}}\big)$
& $0.01$--$0.02$ \\
\midrule
\textbf{Total} & Offline + Online
& $\Theta\!\big(\ell\, d_{\mathrm{sub}}\, n_1^2\big)$
& $< 0.03$ \\
\bottomrule
\end{tabular}
\end{table}
Here, $n_1$ denotes the size of each subgroup, obtained by partitioning the $n$ users into $\ell$ groups ($n_1 = n/\ell$), and $F_{\mathrm{sub}}$ denotes the majority vote polynomial evaluated within each subgroup with degree $d_{\mathrm{sub}} = \deg(F_{\mathrm{sub}})$. 
As shown in Table~\ref{tab:complexity_time}, the total runtime of Algorithm~1 remains below approximately $0.03$ seconds on average, even under classical Beaver triple generation, where the offline preprocessing cost scales as $\Theta(\ell d_{\mathrm{sub}} n_1^2)$ and the online polynomial evaluation requires only $\Theta(\ell d_{\mathrm{sub}})$ field multiplications. In contrast, Algorithm~2—which performs secure aggregation together with local training and sign processing—typically incurs on the order of $10$ seconds per global round in FL tasks such as FMNIST under non-IID settings. Therefore, the additional overhead introduced by Algorithm~1 is negligible relative to the overall cost of a federated training round and does not form a practical bottleneck in our experiments. Note that in practice, runtime is dominated by implementation overheads, making the reported values conservative upper bounds that can be reduced with low-level optimizations.

\section{Experiments}

\subsection{Experiment Setup}
To evaluate the effectiveness and practicality of the proposed Hi-SAFE framework, we conducted experiments on multiple benchmark datasets, including MNIST~\cite{LeCun1998gradient}, FMNIST~\cite{Xiao2017fashion}, and CIFAR-10~\cite{Krizhevsky2009learning}. The datasets are divided as follows: 60,000 training and 10,000 testing samples for both MNIST and FMNIST, and 50,000 training and 10,000 testing samples for CIFAR-10. We consider $N = 100$ users, each having the same number of data samples. Following the setting in~\cite{Mcmahan2017}, two classes are randomly assigned to each user to induce non-IID data distributions. Batch normalization layers are omitted during training on CIFAR-10. At each global round, a fraction $C$ of users is randomly selected to participate, with $C$ chosen between $0.12$ and $0.36$.

To ensure the statistical validity and reproducibility of our experimental results, we report the mean performance over three independent trials with distinct random seeds, which account for variability due to random initialization and stochastic data sampling. All experiments are conducted on a GPU server equipped with two NVIDIA RTX~3090 GPUs. The main hyperparameter settings for Hi-SAFE are summarized in Table~\ref{table_hyper_both}.

\begin{table}[!htb]
\centering
\setlength{\tabcolsep}{3.5pt}
\caption{Hyperparameters for Hi-SAFE}
\label{table_hyper_both}
\begin{tabular}{ |c|c|c|c|c|} 
 \hline
Method & Dataset & $\eta$ (learning rate) & Batch Size & Local Epoch \\
 \hline
\multirow{3}{5.5em}{Non-Subgrouping}  
 & MNIST    & $0.001$  & $100$ & $1$ \\ 
 \cline{2-5}
 & FMNIST   & $0.005$  & $100$ & $1$ \\
 \cline{2-5}
 & CIFAR-10 & $0.0001$ & $100$ & $1$ \\
 \hline 
\multirow{3}{5.5em}{Subgrouping}  
 & MNIST    & $0.001$  & $100$ & $1$ \\
 \cline{2-5}
 & FMNIST   & $0.005$  & $100$ & $1$ \\
 \cline{2-5}
 & CIFAR-10 & $0.0001$ & $100$ & $1$ \\
 \hline 
\end{tabular}
\end{table}

\subsection{Experiment Results}
\begin{figure}[!b]
     \centering
     \begin{subfigure}[b]{0.48\textwidth}
         \centering
         \includegraphics[width=\textwidth]{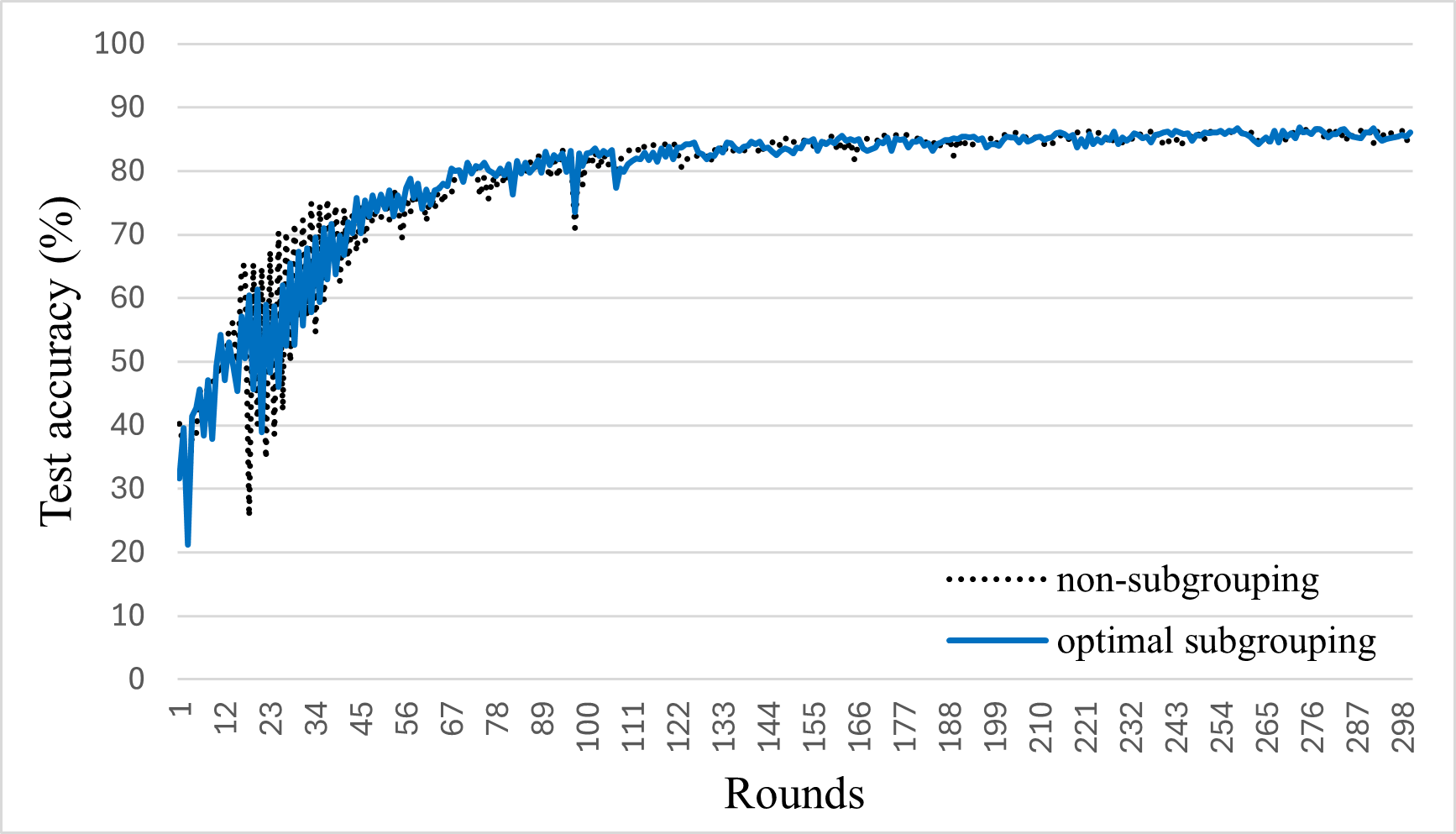}
         \caption{1-bit tie-breaking ($\text{sign}(0)\in \{-1, +1\}$)}
         \label{fig:performance_Case_A_1}
     \end{subfigure}
     \hfill
     \begin{subfigure}[b]{0.48\textwidth}
         \centering
         \includegraphics[width=\textwidth]{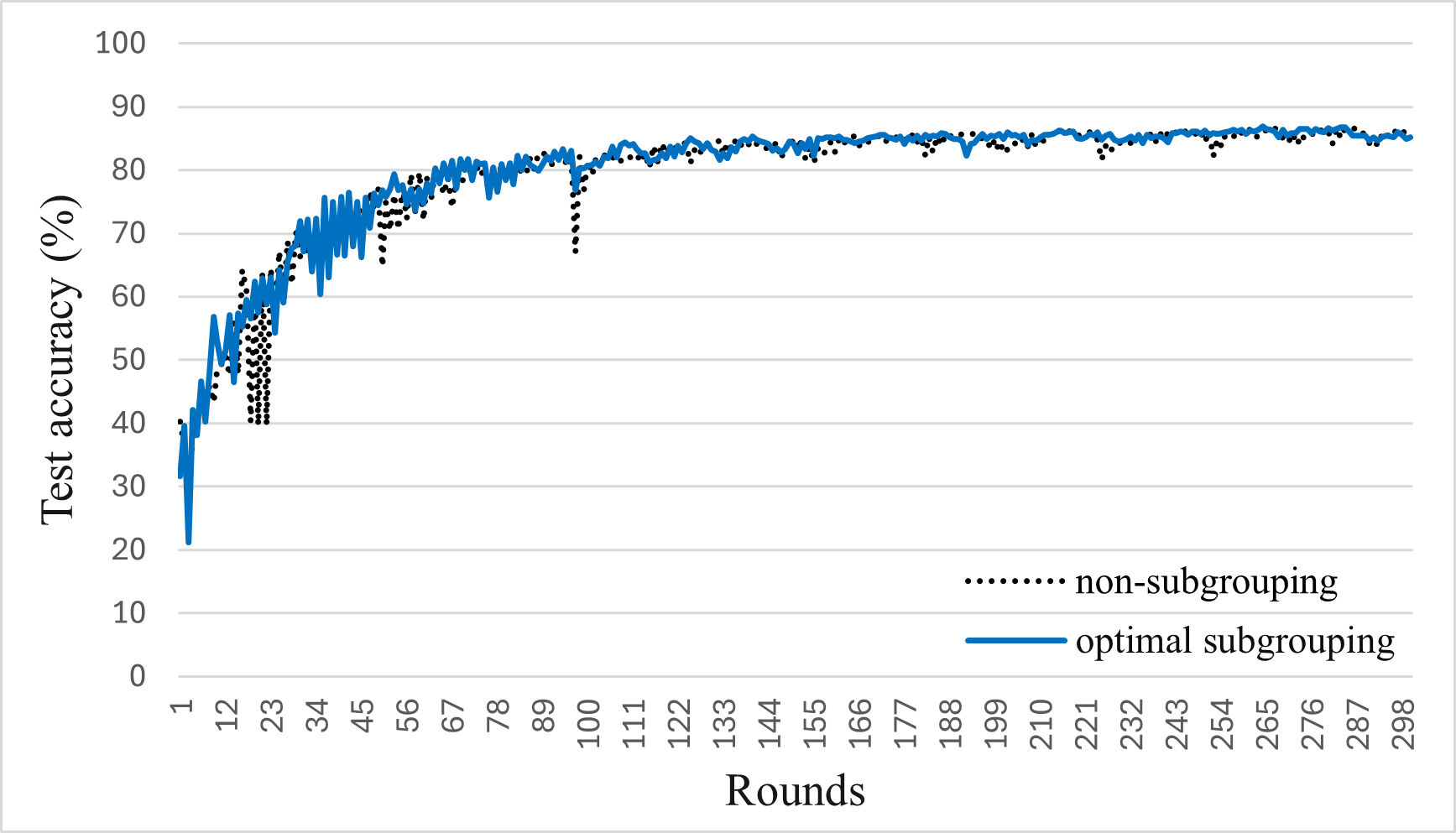}
         \caption{2-bit tie-breaking ($\text{sign}(0)=0$)}
         \label{fig:performance_Case_B_1}
     \end{subfigure}
     \caption{Performance comparison of different tie-breaking policies on the 
     FMNIST dataset with $n = 24$.}
     \label{fig:BTSA_subg_fmnist}
\end{figure}

We first investigate the impact of tie-breaking policies on the performance of Hi-SAFE under both non-subgrouping and optimal subgrouping. Fig.~\ref{fig:BTSA_subg_fmnist} shows representative results on the FMNIST dataset with $n = 24$ users. Fig.~\ref{fig:performance_Case_A_1} corresponds to the baseline configuration in which 1-bit tie-breaking is applied to both intra- and inter-subgroup aggregation, while Fig.~\ref{fig:performance_Case_B_1} applies 2-bit tie-breaking only to intra-subgroup aggregation and retains a 1-bit global update.

The experimental results indicate that both 1-bit and 2-bit tie-breaking strategies yield comparable model accuracy, each exhibiting distinct trade-offs. While 1-bit tie-breaking reduces computational complexity, 2-bit tie-breaking slightly reduces the number of terms in $F(\xv)$ and improves numerical precision. Since all intra-subgroup computations are executed entirely on the server side, this refinement does not introduce any additional uplink communication cost (see Section~\ref{sec:tie-breaking policy}), although it increases server-side computational load. Accordingly, the choice between the two strategies should be guided by the desired balance between computational efficiency and model accuracy.

\begin{figure}[!tb]
     \centering
     \begin{subfigure}[b]{0.48\textwidth} %0.48
         \centering
         \includegraphics[width=\textwidth]{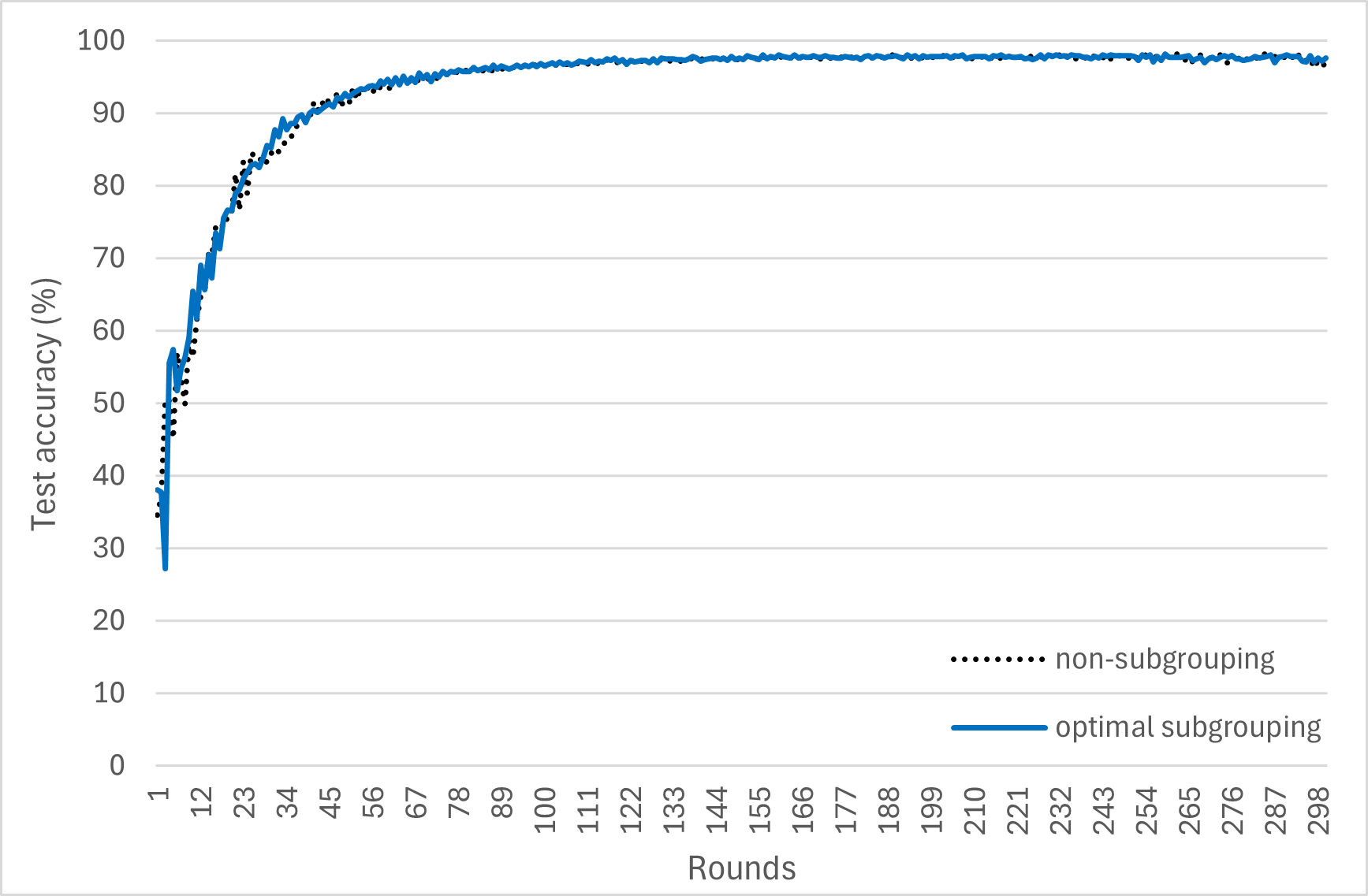}
         \caption{1-bit tie-breaking (Case A-1)}
         %\caption{$\text{sign}(0)\in \{-1, +1\})$ (Case A-1)}
         \label{fig:performance_12_mnist_A-1}
     \end{subfigure}
     \hfill
     \begin{subfigure}[b]{0.48\textwidth}
         \centering
         \includegraphics[width=\textwidth]{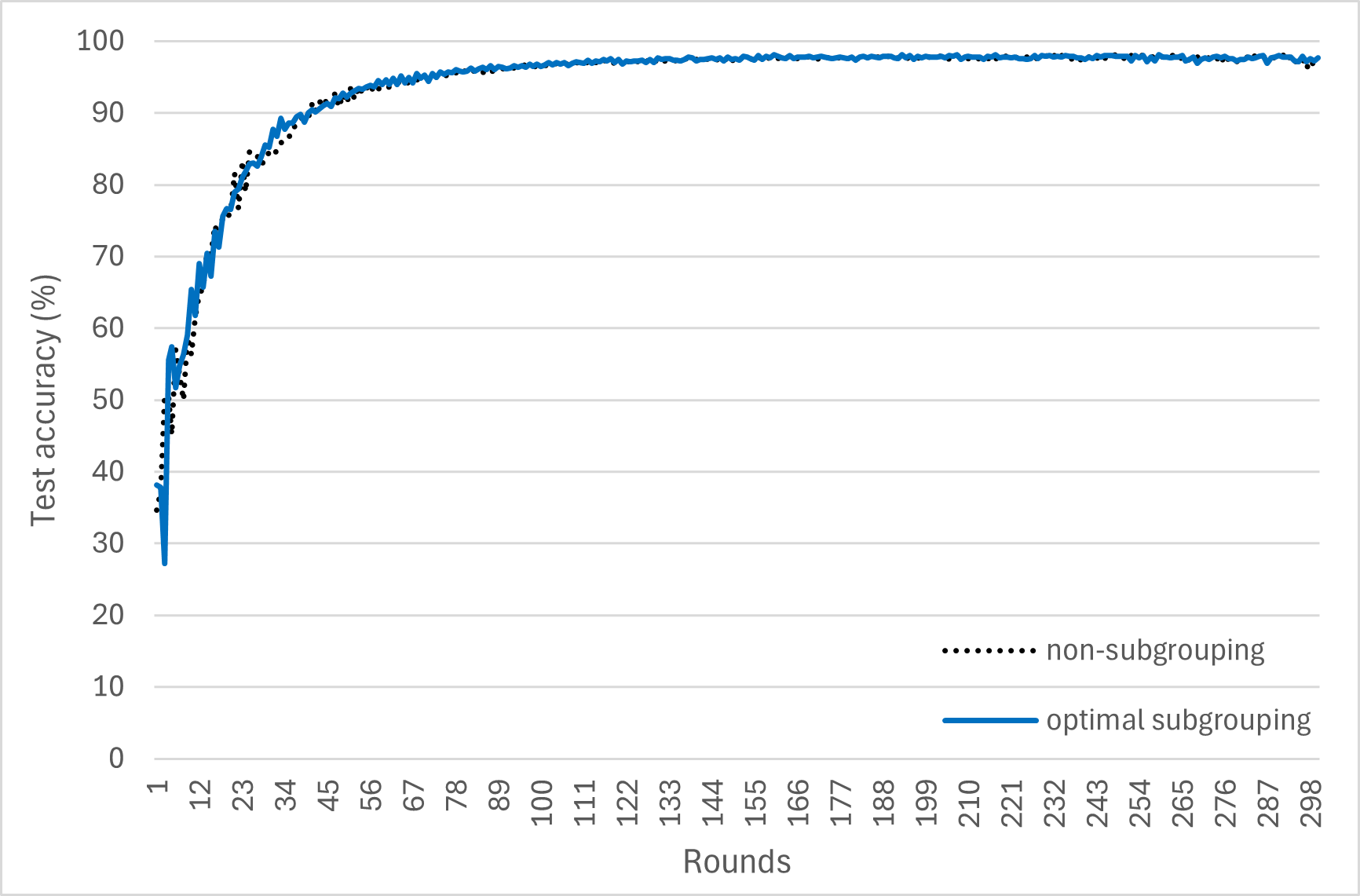}
         \caption{2-bit tie-breaking (Case B-1)}
         %\caption{$\text{sign}(0)=0$ (Case B-1)} 
         \label{fig:performance_12_mnist_B-1}
     \end{subfigure}
        %\caption{\!Performance comparison of different tie-breaking policies on the MNIST dataset with \(\!n \!=\! 12\) and IID.}
        \caption{Performance comparison of tie-breaking policies on the MNIST dataset under IID setting with $n \!=\! 12$.}
        \label{fig:BTSA_subg_12_mnist}
\end{figure}

\begin{figure}[!tb]
     \centering
     \begin{subfigure}[b]{0.48\textwidth}
         \centering
         \includegraphics[width=\textwidth]{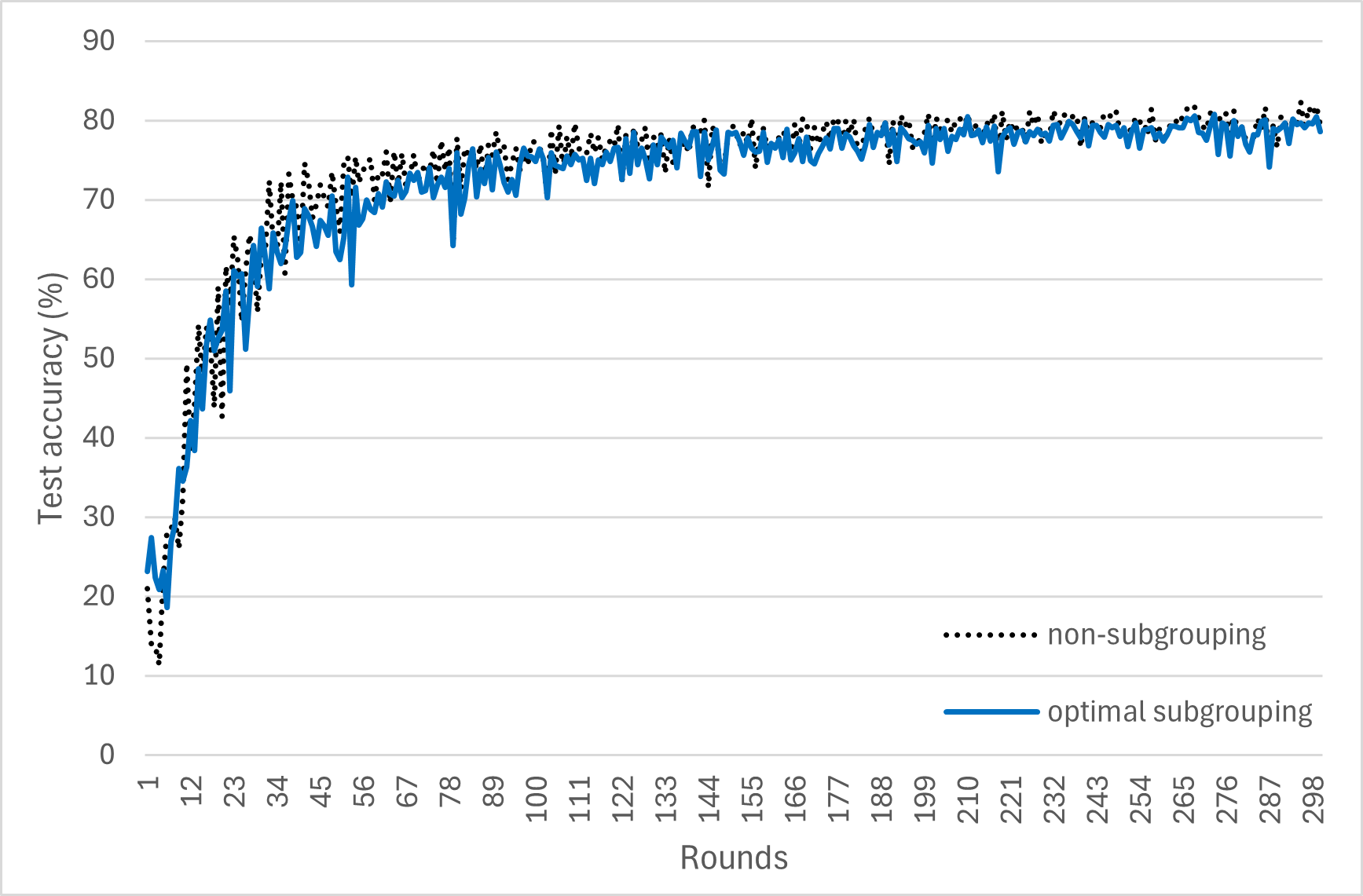}
         \caption{1-bit tie-breaking (Case A-1)}
         %\caption{$\text{sign}(0)\in \{-1, +1\})$ (Case A-1)}
         \label{fig:performance_24_fmnist_nonIID_A-1}
     \end{subfigure}
     \hfill
     \begin{subfigure}[b]{0.48\textwidth}
         \centering
         \includegraphics[width=\textwidth]{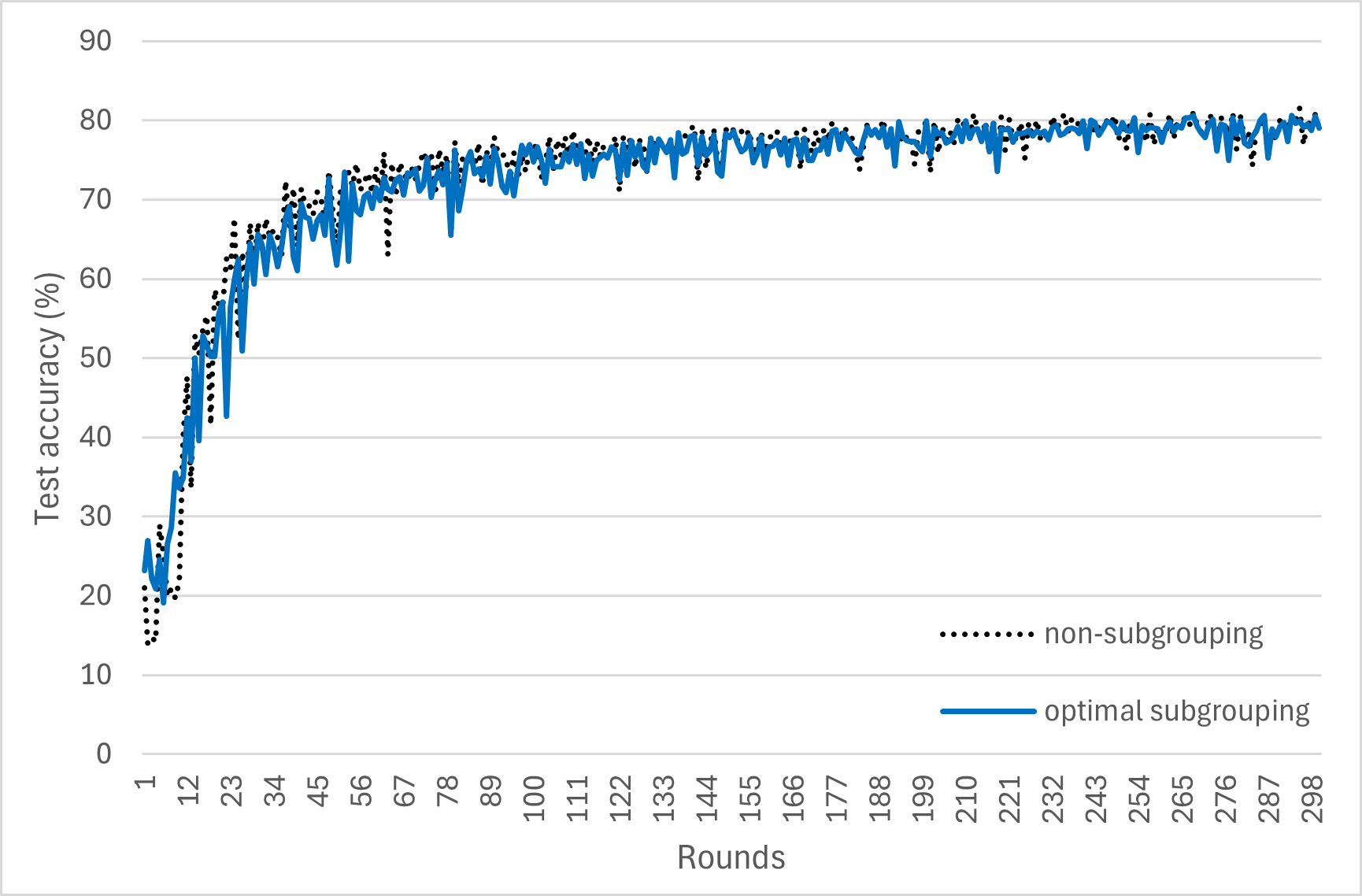}
         \caption{2-bit tie-breaking (Case B-1)}
         %\caption{$\text{sign}(0)=0$ (Case B-1)}
         \label{fig:performance_24_fmnist_nonIID_B-1}
     \end{subfigure}
        %\caption{\!Performance comparison of different tie-breaking policies on the FMNIST dataset with \(\!n \!=\! 24\) and non-IID.}
        \caption{Performance comparison of tie-breaking policies on the FMNIST dataset under non-IID setting with $n \!=\! 24$.}
        \label{fig:BTSA_subg_24_fmnist_nonIID}
\end{figure}

\begin{figure}[!tb]
     \centering
     \begin{subfigure}[b]{0.48\textwidth}
         \centering
         \includegraphics[width=\textwidth]{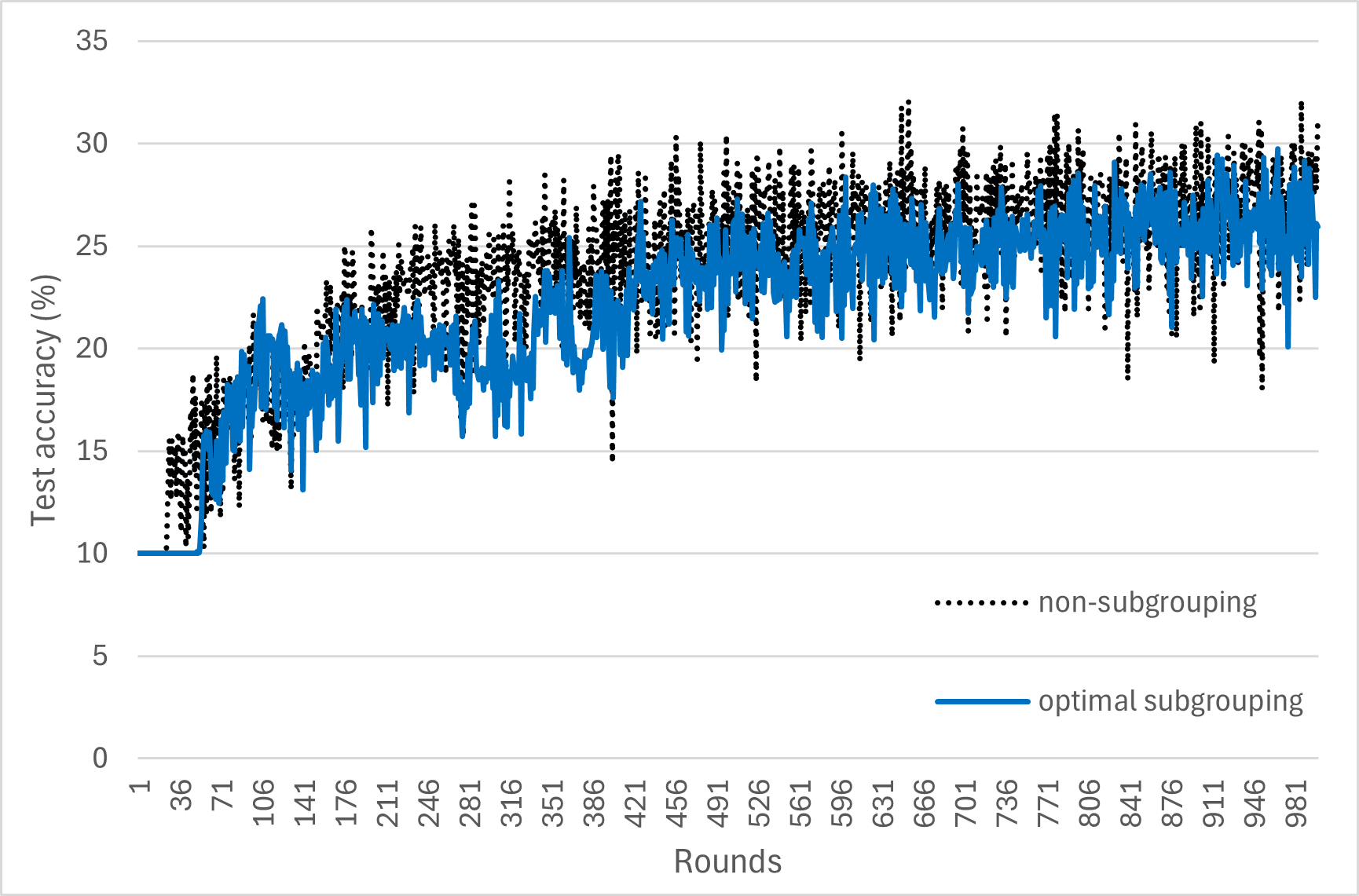}
         \caption{1-bit tie-breaking (Case A-1)}
         %\caption{$\text{sign}(0)\in \{-1, +1\})$ (Case A-1)}
         \label{fig:performance_24_cifar_nonIID_A-1}
     \end{subfigure}
     \hfill
     \begin{subfigure}[b]{0.48\textwidth}
         \centering
         \includegraphics[width=\textwidth]{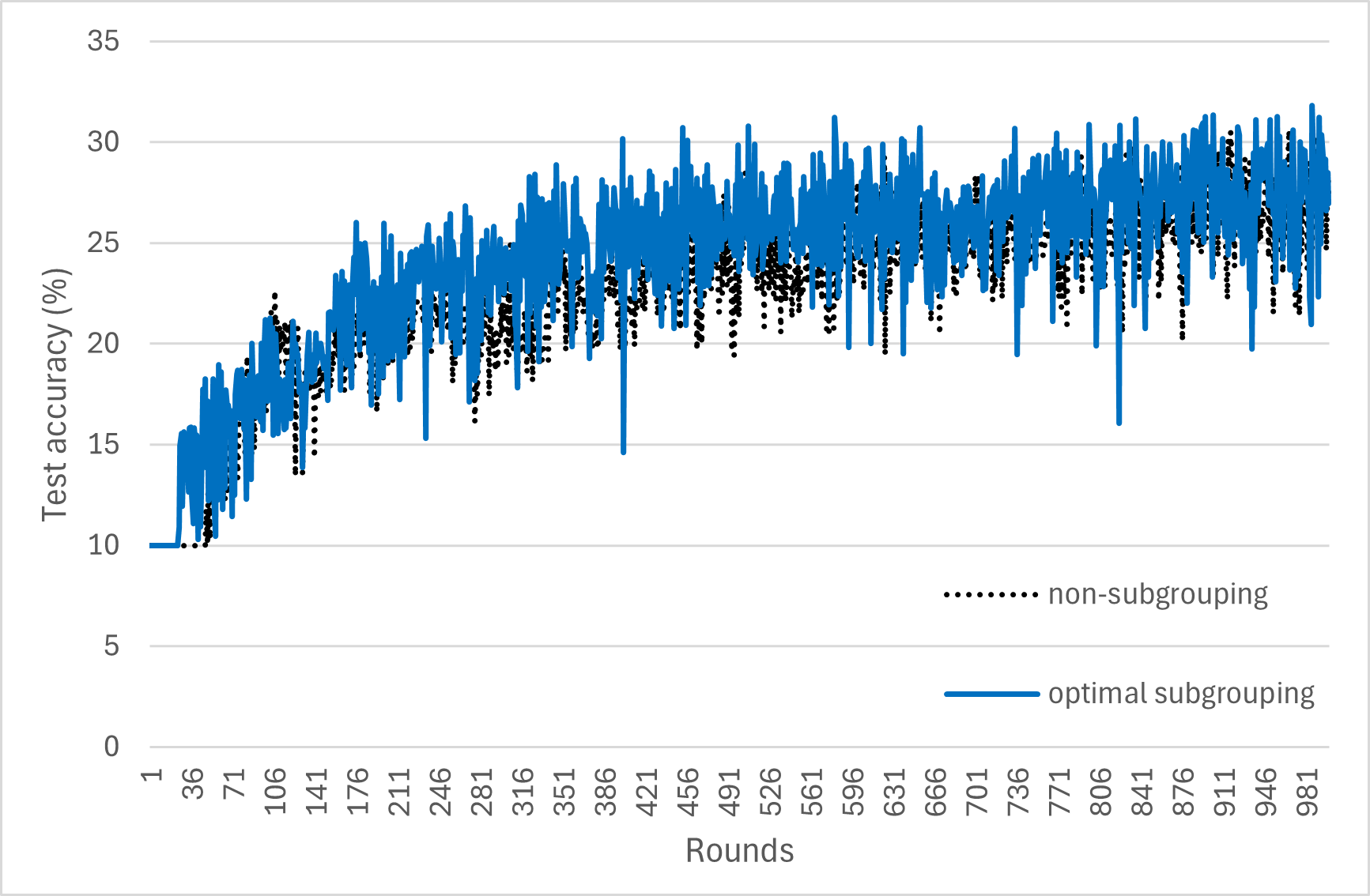}
         \caption{2-bit tie-breaking (Case B-1)}
         %\caption{$\text{sign}(0)=0$ (Case B-1)}
         \label{fig:performance_24_cifar_nonIID_B-1}
     \end{subfigure}
        %\caption{\!Performance comparison of different tie-breaking policies on the FMNIST dataset with \(\!n \!=\! 24\) and non-IID.}
        \caption{Performance comparison of tie-breaking policies on the CIFAR-10 dataset under non-IID setting with $n \!=\! 24$.}
        \label{fig:BTSA_subg_24_cifar_nonIID}
\end{figure}

We further validate these observations across a broad range of FL environments, including MNIST, FMNIST, and CIFAR-10 under both IID and non-IID data settings and for user scales, as illustrated in Figs~\ref{fig:BTSA_subg_12_mnist}–\ref{fig:BTSA_subg_24_cifar_nonIID}. Across all these configurations, the proposed subgrouping strategy maintains model accuracy comparable to the flat (non-subgrouping) setting while significantly reducing the cost of secure computation. Under the 1-bit tie-breaking setting, the non-subgrouping configuration of Hi-SAFE is functionally equivalent to naive \textsc{signSGD-MV}, except for its privacy guarantees, and achieves similar performance to existing methods. In contrast, applying 2-bit tie-breaking improves computational precision on the server side and leads to consistent, though modest, accuracy gains, with the improvements being more pronounced in heterogeneous or challenging tasks (e.g., CIFAR-10, non-IID data setting).

% ----------- appendix 그림
% IID MNIST
%n=12

% non-IID FMNIST 

Overall, these results demonstrate that Hi-SAFE preserves model performance across diverse datasets, data distributions, and user scales, while enabling flexible trade-offs between computational efficiency and accuracy through the choice of tie-breaking policy. This accuracy robustness underpins the subgrouping analysis in the following subsection, where we examine communication and latency benefits.

% ==========================================================

\subsection{Evaluation of Optimal Subgrouping Strategy}
To evaluate the impact of subgrouping on communication efficiency, we characterize the 
total communication cost \(C_T\) and the per-user communication cost \(C_u\). Let 
\(\ell = n / n_1\) denote the number of subgroups, where \(\ell = 1\) corresponds to the 
non-subgrouping case. Here, \(n\) is the total number of users and \(n_1\) denotes the number 
of users assigned to each subgroup. Let \(p_1\) be the smallest prime strictly greater than 
\(n_1\), and let \(\lceil \log p_1 \rceil\) denote its bit length used for field representation. 
The secure multiplication procedure requires \(\lceil \log p_1 - 1 \rceil\) sequential Beaver 
subrounds, and the total number of required secure multiplications is \(R\), which grows 
proportionally with this multiplicative depth. Under these definitions, the total 
communication cost and the per-user cost are given by
\[
C_T = \ell \cdot \big(R \cdot \lceil \log p_1 \rceil\big), \qquad
C_u = R \cdot \lceil \log p_1 \rceil \quad \text{(bits)}.
\]

Table~\ref{table:extended_comm_cost} summarizes the optimal subgroup configurations $\ell^\star$ that minimize the total communication cost $C_T$ for various user counts $n$. Parentheses denote the percentage reduction relative to the non-subgrouping baseline. These results demonstrate that Hi-SAFE achieves substantial reductions in both total and per-user communication costs ($C_u$) without degrading model accuracy. Notably, for $n \geq 24$, the per-user communication cost consistently decreases by more than $94\%$, with up to $52.0\%$ reduction in total communication cost observed at $n = 24$. These findings validate the scalability and communication efficiency of the proposed framework.

\begin{table}[!t]
\centering
\setlength{\tabcolsep}{3.3pt}
\caption{Optimal subgroup configuration and communication cost}
\label{table:extended_comm_cost}
\begin{tabular}{|c|c|c|c|c|c|c|}
\hline
$n$ & $\ell^\star$ & $n_1$ & $\lceil \log p_1 \!-\! 1 \rceil$ & \#multiplications & $C_T$ (\%) & $C_u$ (\%) \\
\hline
$24$  & $8$  & $3$  & $2$ & $4$  & $96$ (52.0\%)   & $12$ (94.0\%)   \\
$36$  & $12$ & $3$  & $2$ & $4$  & $144$ (47.8\%)  & $12$ (95.7\%)   \\
$60$  & $20$ & $3$  & $2$ & $4$  & $240$ (44.4\%)  & $12$ (97.2\%)   \\
$90$  & $30$ & $3$  & $2$ & $4$  & $360$ (50.5\%)  & $12$ (98.4\%)   \\
$100$ & $25$ & $4$  & $2$ & $6$  & $450$ (43.6\%)  & $18$ (97.7\%)   \\
\hline
\end{tabular}
\end{table}

\begin{figure}[!t]
    \centering
    \begin{subfigure}[b]{0.49\textwidth}
        \centering
        \includegraphics[width=\textwidth]{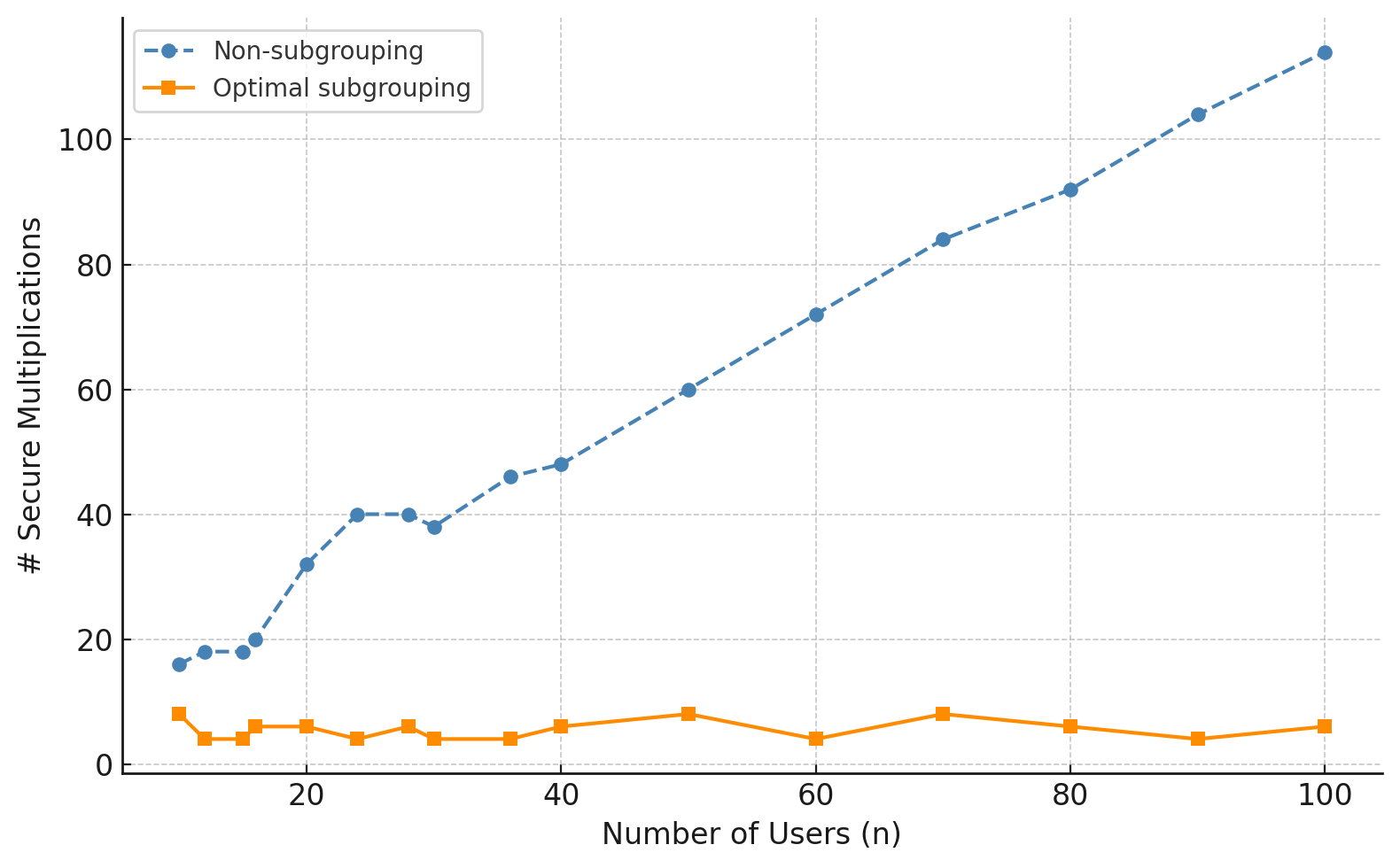}
        \caption{Per-user secure multiplications}
        \label{fig:secure_multiplication_shared}
    \end{subfigure}
    \hfill
    \begin{subfigure}[b]{0.49\textwidth}
        \centering
        \includegraphics[width=\textwidth]{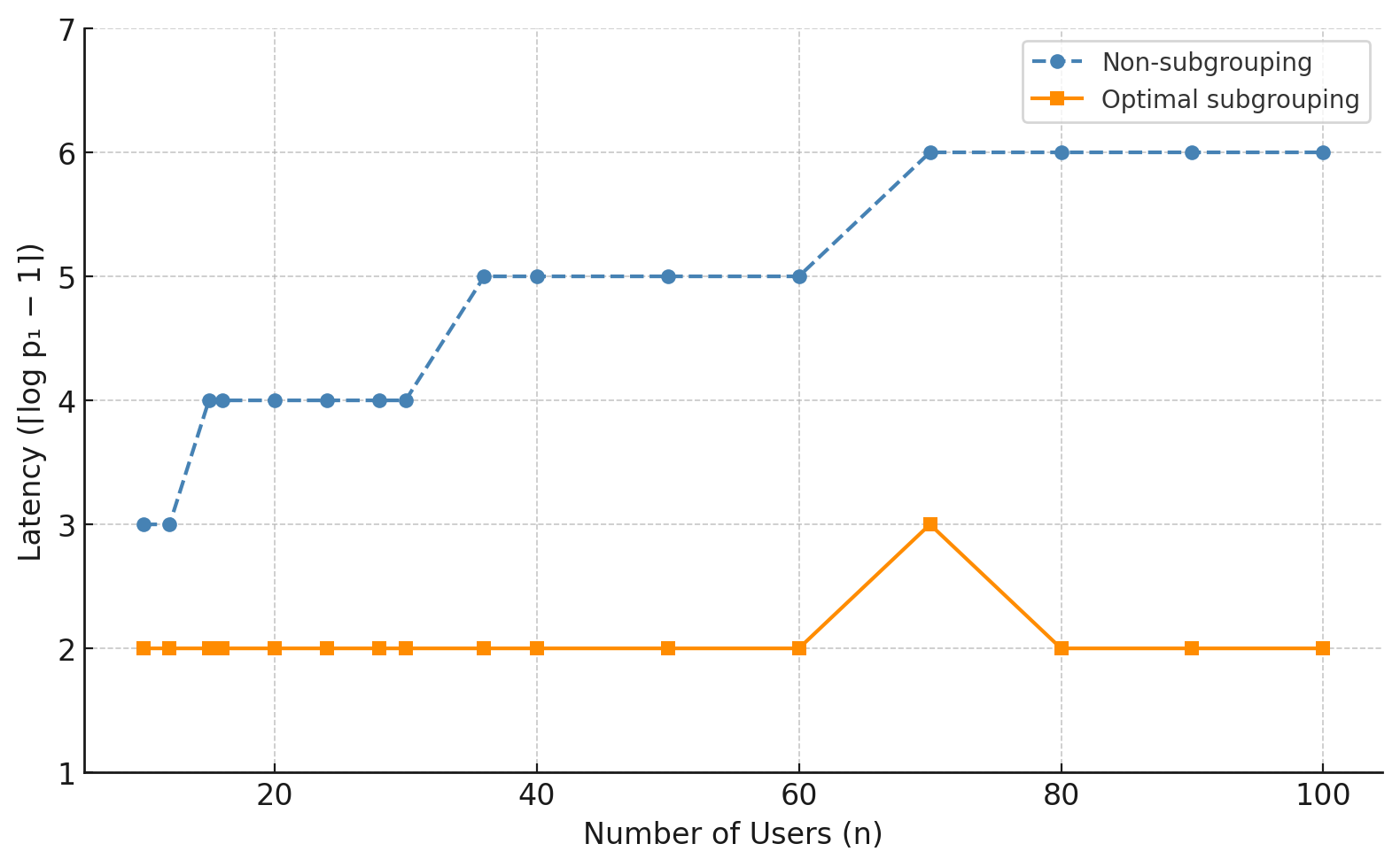}
        \caption{Latency $\lceil \log p_1 \!-\! 1 \rceil$ for secure multiplication}
        \label{fig:latency_comparison}
    \end{subfigure}
    \caption{Impact of optimal subgrouping on secure multiplication cost and latency.}
    \label{fig:subgrouping_analysis}
\end{figure}

Fig.~\ref{fig:subgrouping_analysis} further illustrates the effect of optimal subgrouping on per-user secure multiplications and their latency. In the non-subgrouping setting (Fig.~\ref{fig:secure_multiplication_shared}), a global majority vote polynomial must be evaluated over all $n$ users, resulting in a per-user cost that grows linearly with $n$. In contrast, the proposed subgrouping strategy partitions users into subgroups of size $n_1$, enabling each group to evaluate a much smaller majority vote polynomial. This keeps the per-user cost constant and low ($\le 6$ multiplications), regardless of system scale.

Fig.~\ref{fig:latency_comparison} shows the latency, defined as the serial depth $\lceil \log p_1 - 1 \rceil$ for Beaver triple multiplication. In the non-subgrouping case, larger finite fields are required to support the global majority vote polynomial, which leads to increased latency. Subgrouping, on the other hand, confines computation to smaller groups, enabling the use of smaller fields and consistently achieving low latency—often as low as $2$. These results confirm that optimal subgrouping yields substantial communication and latency savings while preserving the accuracy guarantees demonstrated in the previous subsection. %To complement these findings, we further evaluate the effect of varying subgrouping configurations, parameterized by \(\ell\), in Section~\ref{sec:additional_evaluation}.

%%%%%%%%% subgrouping \ell^*와 통신량 분석
\subsubsection{Effect of the Subgrouping Configuration} 
\label{sec:additional_evaluation}
To complement these findings, we further evaluate the effect of varying subgrouping configurations. In particular, we assess how varying the subgroup size influences the overall communication cost, per-user overhead, latency, and multiplication complexity.

Tables~\ref{table:comm_cost} and \ref{table:comm_cost2} present a comparison of the per-user communication cost \(C_u\) and total cost \(C_T\), as well as the associated latency and multiplication cost under various subgroup configurations. Parentheses in the table denote the percentage reduction relative to the baseline case \(\ell = 1\), that is, the non-subgrouping case. The results show that choosing an optimal \(\ell\) leads to substantial savings not only in total and per-user communication costs but also in the associated latency and multiplication cost, thereby validating the effectiveness of the proposed subgrouping strategy for scalable and communication-efficient secure aggregation in FL.

\begin{table}[!htb]
\centering
\setlength{\tabcolsep}{2.7pt}
\caption{Key metrics across different subgroup configurations}
\label{table:comm_cost}
\begin{tabular}{|c|c|c|c|c|c|c|c|c|}
\hline
$n$ & $\ell$ & $n_1$ & $p_1$ & $\lceil \log p_1 \rceil$ & $\lceil \log p_1 \!-\! 1 \rceil$ & $R$ & $C_T$ (\%) & $C_u$ (\%) \\
\hline
%10 & 1 & 10 & 11 & 4 & 3 & 16 & 64 (-) & 64 (-) \\
%10 & 2 & 5  & 7  & 3 & 2 & 8  & 48 (25.0\%) & 24 (62.5\%) \\
%\hline
12 & 1 & 12 & 13 & 4 & 3 & 18 & 72 (-) & 72 (-) \\
12 & 2 & 6  & 7  & 3 & 2 & 10 & 60 (16.7\%) & 30 (58.3\%) \\
12 & 3 & 4  & 5  & 3 & 2 & 6  & 54 (25.0\%) & 18 (75.0\%) \\
12 & 4 & 3  & 5  & 3 & 2 & 4  & 48 (33.3\%) & 12 (83.3\%) \\
\hline
15 & 1 & 15 & 17 & 5 & 4 & 18 & 90 (-) & 90 (-) \\
15 & 3 & 5  & 7  & 3 & 2 & 8  & 48 (46.7\%) & 24 (73.3\%) \\
15 & 5 & 3  & 5  & 3 & 2 & 4  & 60 (33.3\%) & 12 (86.7\%) \\
\hline
16 & 1 & 16 & 17 & 5 & 4 & 20 & 100 (-) & 100 (-) \\
16 & 2 & 8  & 11 & 4 & 3 & 14 & 112 (-12.0\%) & 56 (44.0\%) \\
16 & 4 & 4  & 5  & 3 & 2 & 6  & 72 (28.0\%) & 18 (82.0\%) \\
\hline
20 & 1 & 20 & 23 & 5 & 4 & 32 & 160 (-) & 160 (-) \\
20 & 2 & 10 & 11 & 4 & 3 & 16 & 128 (20.0\%) & 64 (60.0\%) \\
20 & 4 & 5  & 7  & 3 & 2 & 8  & 96 (40.0\%) & 24 (85.0\%) \\
20 & 5 & 4  & 5  & 3 & 2 & 6  & 90 (43.8\%) & 18 (88.7\%) \\
\hline
24 & 1 & 24 & 29 & 5 & 4 & 40 & 200 (-) & 200 (-) \\
24 & 2 & 12 & 13 & 4 & 3 & 18 & 144 (28.0\%) & 72 (64.0\%) \\
24 & 3 & 8  & 11 & 4 & 3 & 14 & 168 (16.0\%) & 56 (72.0\%) \\
24 & 4 & 6  & 7  & 3 & 2 & 10 & 120 (40.0\%) & 30 (85.0\%) \\
24 & 6 & 4  & 7  & 3 & 2 & 6  & 108 (46.0\%) & 18 (91.0\%) \\
24 & 8 & 3  & 5  & 3 & 2 & 4  & 96 (52.0\%) & 12 (94.0\%) \\
\hline
28 & 1 & 28 & 29 & 5 & 4 & 40 & 200 (-) & 200 (-) \\
28 & 2 & 14 & 17 & 5 & 4 & 22 & 220 (-10.0\%) & 110 (45.0\%) \\
28 & 4 & 7  & 11 & 4 & 3 & 14 & 224 (-12.0\%) & 56 (72.0\%) \\
28 & 7 & 4  & 5  & 3 & 2 & 6  & 126 (37.0\%) & 18 (91.0\%) \\
\hline
30 & 1 & 30 & 31 & 5 & 4 & 38 & 190 (-) & 190 (-) \\
30 & 2 & 15 & 17 & 4 & 3 & 20 & 200 (-5.3\%) & 100 (47.4\%) \\
30 & 3 & 10 & 11 & 4 & 3 & 16 & 192 (-1.1\%) & 64 (66.3\%) \\
30 & 5 & 6  & 7  & 3 & 2 & 10 & 150 (21.1\%) & 30 (84.2\%) \\
30 & 6 & 5  & 7  & 3 & 2 & 8  & 144 (24.2\%) & 24 (87.4\%) \\
30 & 10 & 3 & 5  & 3 & 2 & 4  & 120 (36.8\%) & 12 (93.7\%) \\
\hline
36 & 1 & 36 & 37 & 6 & 5 & 46 & 276 (-) & 276 (-) \\
36 & 2 & 18 & 19 & 5 & 4 & 26 & 260 (5.8\%) & 130 (52.9\%) \\
36 & 3 & 12 & 13 & 4 & 3 & 18 & 216 (21.7\%) & 72 (73.9\%) \\
36 & 4 & 9  & 11 & 4 & 3 & 14 & 224 (18.8\%) & 56 (79.7\%) \\
36 & 6 & 6  & 7  & 3 & 2 & 10 & 180 (34.8\%) & 30 (89.1\%) \\
36 & 9 & 4  & 5  & 3 & 2 & 6  & 162 (41.3\%) & 18 (93.5\%) \\
36 & 12 & 3 & 5  & 3 & 2 & 4  & 144 (47.8\%) & 12 (95.7\%) \\
\hline
40 & 1 & 40 & 41 & 6 & 5 & 48 & 288 (-) & 288 (-) \\
40 & 2 & 20 & 23 & 5 & 4 & 32 & 320 (-11.1\%) & 160 (44.4\%) \\
40 & 4 & 10 & 11 & 4 & 3 & 16 & 256 (11.1\%) & 64 (77.8\%) \\
40 & 5 & 8  & 11 & 4 & 3 & 14 & 280 (2.8\%) & 56 (80.6\%) \\
40 & 8 & 5  & 7  & 3 & 2 & 8  & 192 (33.3\%) & 24 (91.7\%) \\
40 & 10& 4  & 5  & 3 & 2 & 6  & 180 (37.5\%) & 18 (93.8\%) \\
\hline
\end{tabular}
\end{table}

\begin{table}[!htb]
\centering
\setlength{\tabcolsep}{2.3pt}
\caption{Key metrics across different subgroup configurations (continue)}
\label{table:comm_cost2}
\begin{tabular}{|c|c|c|c|c|c|c|c|c|}
\hline
$n$ & $\ell$ & $n_1$ & $p_1$ & $\lceil \log p_1 \rceil$ & $\lceil \log p_1 - 1 \rceil$ & $R$ & $C_T$ (\%) & $C_u$ (\%) \\
\hline
50 & 1 & 50 & 51 & 6 & 5 & 60 & 360 (-) & 360 (-) \\
50 & 2 & 25 & 29 & 5 & 4 & 34 & 340 (5.6\%) & 170 (52.8\%) \\
50 & 5 & 10 & 11 & 4 & 3 & 16 & 320 (11.1\%) & 64 (82.2\%) \\
50 & 10& 5  & 7  & 3 & 2 & 8  & 240 (33.3\%) & 24 (93.3\%) \\
\hline
60 & 1 & 60 & 61 & 6 & 5 & 72 & 432 (-) & 432 (-) \\
60 & 2 & 30 & 31 & 5 & 4 & 38 & 380 (12.0\%) & 190 (56.0\%) \\
60 & 3 & 20 & 23 & 5 & 3 & 32 & 480 (-11.1\%) & 160 (63.0\%) \\
60 & 5 & 12 & 13 & 4 & 3 & 18 & 360 (16.7\%) & 72 (83.3\%) \\
60 & 6 & 10 & 11 & 4 & 2 & 16 & 384 (11.1\%) & 64 (85.2\%) \\
60 & 10& 6  & 7  & 3 & 2 & 10 & 300 (30.6\%) & 30 (93.1\%) \\
60 & 12& 5  & 7  & 3 & 2 & 8  & 288 (33.3\%) & 24 (94.4\%) \\
60 & 20& 3  & 5  & 3 & 2 & 4  & 240 (44.4\%) & 12 (97.2\%) \\
\hline
70 & 1 & 70 & 71 & 7 & 6 & 84 & 588 (-) & 588 (-) \\
70 & 2 & 35 & 37 & 6 & 5 & 44 & 528 (10.2\%) & 264 (55.1\%) \\
70 & 5 & 14 & 17 & 5 & 4 & 22 & 550 (6.5\%)  & 110 (81.3\%) \\
70 & 7 & 10 & 11 & 4 & 3 & 16 & 448 (23.8\%) & 64 (89.1\%) \\
70 & 10& 7  & 11 & 4 & 3 & 14 & 560 (4.8\%)  & 56 (90.5\%) \\
70 & 14& 5  & 7  & 3 & 3 & 8  & 336 (42.9\%) & 24 (95.9\%) \\
\hline
80 & 1 & 80 & 81 & 7 & 6 & 92 & 644 (-) & 644 (-) \\
80 & 2 & 40 & 41 & 6 & 5 & 48 & 576 (10.6\%) & 288 (55.3\%) \\
80 & 4 & 20 & 23 & 5 & 4 & 32 & 640 (0.6\%)  & 160 (75.2\%) \\
80 & 5 & 16 & 17 & 5 & 4 & 20 & 500 (22.4\%) & 100 (84.5\%) \\
80 & 8 & 10 & 11 & 4 & 3 & 16 & 512 (20.6\%) & 64 (90.1\%) \\
80 & 10& 8  & 11 & 4 & 3 & 14 & 560 (13.0\%) & 56 (91.3\%) \\
80 & 16& 5  & 7  & 3 & 2 & 8  & 384 (40.4\%) & 24 (96.3\%) \\
80 & 20& 4  & 5  & 3 & 2 & 6  & 360 (44.1\%) & 18 (97.2\%) \\
\hline
90 & 1 & 90 & 91 & 7 & 6 & 104& 728 (-) & 728 (-) \\
90 & 2 & 45 & 47 & 6 & 5 & 54 & 648 (11.0\%) & 324 (55.5\%) \\
90 & 3 & 30 & 31 & 5 & 4 & 38 & 570 (21.7\%) & 190 (73.9\%) \\
90 & 5 & 18 & 19 & 5 & 4 & 26 & 650 (10.7\%) & 130 (82.1\%) \\
90 & 6 & 15 & 17 & 5 & 4 & 18 & 540 (25.8\%) & 90 (87.6\%) \\
90 & 9 & 10 & 11 & 4 & 3 & 16 & 576 (20.9\%) & 64 (91.2\%) \\
90 & 10& 9  & 11 & 4 & 3 & 14 & 560 (23.1\%) & 56 (92.3\%) \\
90 & 15& 6  & 7  & 3 & 2 & 10 & 450 (38.2\%) & 30 (95.9\%) \\
90 & 18& 5  & 7  & 3 & 2 & 8  & 432 (40.7\%) & 24 (96.7\%) \\
90 & 30& 3  & 5  & 3 & 2 & 4  & 360 (50.5\%) & 12 (98.4\%) \\
\hline
100 & 1 & 100& 101& 7 & 6 & 114& 798 (-) & 798 (-) \\
100 & 2 & 50 & 51 & 6 & 5 & 60 & 720 (9.8\%)  & 360 (54.9\%) \\
100 & 4 & 25 & 29 & 5 & 4 & 34 & 680 (14.8\%) & 170 (78.7\%) \\
100 & 5 & 20 & 23 & 5 & 4 & 32 & 800 (-0.3\%) & 160 (79.9\%) \\
100 & 10& 10 & 11 & 4 & 3 & 16 & 640 (19.8\%) & 64 (92.0\%) \\
100 & 20& 5  & 7  & 3 & 2 & 8  & 480 (39.9\%) & 24 (97.0\%) \\
100 & 25& 4  & 5  & 3 & 2 & 6  & 450 (43.6\%) & 18 (97.7\%) \\
\hline
\end{tabular}
\end{table}
%%\end{comment}

% ======================================================================================
\section{Conclusion}
In this paper, we have proposed Hi-SAFE, a lightweight and cryptographically secure aggregation framework for communication-efficient and privacy-preserving FL. By securely evaluating majority vote polynomials under additive secret sharing, instantiated for example via Beaver triples, Hi-SAFE achieves end-to-end privacy for sign-based FL, revealing only the final majority vote to the server under the semi-honest model. Furthermore, the proposed hierarchical subgrouping strategy ensures constant latency and a bounded secure multiplication cost per user, independent of the total number of users.
Extensive theoretical and experimental analyses demonstrate that Hi-SAFE reduces per-user communication cost by over 94\% when \(n \geq 24\), and achieves up to 52\% reduction in total communication cost at \(n = 24\), while preserving model accuracy. These results confirm the scalability, robustness, and practicality of Hi-SAFE, especially in bandwidth-constrained FL deployments.

% ==================================================================================
%
% ================================ Appendix ========================================
\appendices

\section{Illustrative Example: Secure Evaluation of the Majority Vote Polynomial}
\label{sec:toy example}

To aid understanding, we present a step-by-step example of securely computing the majority vote polynomial \( F(\xv) \) using Beaver triples. In particular, we describe the detailed procedure by which each user \(i\) securely evaluates an encrypted share \(\llbracket F(\xv) \rrbracket_i\), which is subsequently transmitted to Server for aggregation.

\subsection{\textbf{Secure Evaluation of $F(\xv) = 2\xv^3 + 4\xv \pmod{5}$ with $n = 3$}}

As a concrete example, consider the evaluation of the polynomial \( F(\xv) = 2\xv^3 + 4\xv \pmod{5} \) over the finite field \(\mathbb{F}_5\), assuming \( n = 3 \) users. Each user holds a private input \( \xv_i \in \{-1, +1\}\) such that:
\[
\xv = \sum_{i=1}^{3} \xv_i.
\]

For simplicity, we assume the following scalar user inputs:
\[
x_1 = 1,\quad x_2 = -1,\quad x_3 = 1,
\]
which yield the majority vote result:
\[
\text{sign} \left( \sum_{i=1}^3 x_i \right) = \text{sign}(1 - 1 + 1) = \text{sign}(1) = 1.
\]

To evaluate \( F(x) \) securely, we adopt a Beaver triple-based protocol. In this example, the following Beaver triples are pre-shared among users during the offline phase:
\[
a^r = \sum_{i=1}^3 \llbracket a^r \rrbracket_i,\, b^r = \sum_{i=1}^3 \llbracket b^r \rrbracket_i,\, c^r = a^r \cdot b^r = \sum_{i=1}^3 \llbracket c^r \rrbracket_i,\, r \in \{1,2\},
\]
with each share lying in the field \(\mathbb{F}_5\). The specific shares are given by:
\[
\begin{aligned}
&\llbracket a^1 \rrbracket_1 = 0,\,\llbracket a^1 \rrbracket_2 = 3,\, \llbracket a^1 \rrbracket_3 = 2,
&&\!\!\!\!\llbracket a^2 \rrbracket_1 = 4,\, \llbracket a^2 \rrbracket_2 = 3,\, \llbracket a^2 \rrbracket_3 = 1, \\
&\llbracket b^1 \rrbracket_1 = 2,\, \llbracket b^1 \rrbracket_2 = 2,\, \llbracket b^1 \rrbracket_3 = 0,
&&\!\!\!\!\llbracket b^2 \rrbracket_1 = 0,\, \llbracket b^2 \rrbracket_2 = 1,\, \llbracket b^2 \rrbracket_3 = 4, \\
&\llbracket c^1 \rrbracket_1 = 1,\, \llbracket c^1 \rrbracket_2 = 1,\, \llbracket c^1 \rrbracket_3 = 3,
&&\!\!\!\!\llbracket c^2 \rrbracket_1 = 1,\, \llbracket c^2 \rrbracket_2 = 2,\, \llbracket c^2 \rrbracket_3 = 2.
\end{aligned}
\]

To evaluate \(F(x)\) securely, it is necessary to first compute the shared cubic term \(x^3\), which can be decomposed as:
\begin{equation}
\begin{aligned}
x^3 &= x \cdot x^2 = (x - a^2 + a^2)(x^2 - b^2 + b^2) \\
&= (x - a^2)(x^2 - b^2) + a^2(x^2 - b^2) + b^2(x - a^2) + c^2 \pmod{5},
\end{aligned}
\label{eq_xxx}
\end{equation}
where \(x^2\) itself is computed via:
\begin{equation}
\begin{aligned}
x^2 &= x \cdot x = (x - a^1 + a^1)(x - b^1 + b^1) \\
&= (x - a^1)(x - b^1) + a^1(x - b^1) + b^1(x - a^1) + c^1 \pmod{5}.
\end{aligned}
\label{eq_xx}
\end{equation}

As the computation of \(x^3\) requires the intermediate value \(x^2\), the evaluation proceeds in two secure multiplication subrounds. We now describe each subround in detail.

%%%%%%%%%%%%%%%%%%%%%%%%%%%%%%%%%%%%%%%%%%%%%%%%%%%%%%%%%%%%%%%%%%%%%%%%%%%%%%%%%%%%%%%%%%%%%%%%%%%%%%%%%%%%%%%%%%%%%%%%%%%%%%%%%%%%%%%%%%%%%%%%%%%%
\begin{itemize}
\item \textbf{In subround 0:} each user \( i \) prepares the necessary values for computing \(x^2\) using Beaver triples. Specifically, each user locally computes the following masked values: % differences = masked values 용어 통일 필요
\[
(x_i - \llbracket a^1 \rrbracket_i) \quad \text{and} \quad (x_i - \llbracket b^1 \rrbracket_i) \pmod{5}.
\]

The local computations for each user are given below:
\begin{align*}
\text{User 1:} \quad & x_1 - \llbracket a^1 \rrbracket_1 = 1 - 0 = 1 \pmod{5}, \\
                     & x_1 - \llbracket b^1 \rrbracket_1 = 1 - 2 = -1 \equiv 4 \pmod{5}, \\
\text{User 2:} \quad & x_2 - \llbracket a^1 \rrbracket_2 = -1 - 3 = -4 \equiv 1 \pmod{5}, \\
                     & x_2 - \llbracket b^1 \rrbracket_2 = -1 - 2 = -3 \equiv 2 \pmod{5}, \\
\text{User 3:} \quad & x_3 - \llbracket a^1 \rrbracket_3 = 1 - 2 = -1 \equiv 4 \pmod{5}, \\
                     & x_3 - \llbracket b^1 \rrbracket_3 = 1 - 0 = 1 \pmod{5}.
\end{align*}

Each user transmits the computed masked values to Server, which then aggregates the results to obtain:
\begin{align*}
x - a^1 &= \sum_{i=1}^{3} (x_i - \llbracket a^1 \rrbracket_i) = 1 + 1 + 4 = 6 \equiv 1\!\! \pmod{5}, \\
x - b^1 &= \sum_{i=1}^{3} (x_i - \llbracket b^1 \rrbracket_i) = 4 + 2 + 1 = 7 \equiv 2\!\! \pmod{5}.
\end{align*}

Server then broadcasts the aggregated values \((x - a^1)\) and \((x - b^1)\) to all users to complete the secure multiplication step for computing \(x^2\).

    \item \textbf{In subround 1:} each user \( i \) prepares the values necessary to compute \(x^3\) using Beaver triples. To this end, the following quantities must be obtained:
\[
(x_i - \llbracket a^2 \rrbracket_i) \quad \text{and} \quad (\llbracket x^2 \rrbracket_i - \llbracket b^2 \rrbracket_i) \pmod{5}.
\]

Here, each user computes \(\llbracket x^2 \rrbracket_i\) based on the pre-shared Beaver triples \((\llbracket a^1 \rrbracket_i, \llbracket b^1 \rrbracket_i, \llbracket c^1 \rrbracket_i)\) and the publicly computable term \((x - a^1)(x - b^1)\), as defined in Eq.~(\ref{eq_xx}):
\[
\llbracket x^2 \rrbracket_i \!=\! (x \!-\! a^1)(x \!-\! b^1) \!+\! \llbracket a^1 \rrbracket_i (x \!-\! b^1) \!+\! \llbracket b^1 \rrbracket_i (x \!-\! a^1) \!+\! \llbracket c^1 \rrbracket_i \!\!\!\pmod{5}.
\]

Since the product \((x - a^1)(x - b^1)\) is constant across all users, only a single user needs to compute and broadcast it to the server. Without loss of generality, we assume that User~1 performs this computation.

Based on the received values, the users compute \(\llbracket x^2 \rrbracket_i\) as follows:
\begin{align*}
\text{User 1:} \, \llbracket x^2 \rrbracket_1 &= 1 \cdot 2 + 0 \cdot 2 + 2 \cdot 1 + 1 = 5 \equiv 0 \pmod{5}, \\
\text{User 2:} \, \llbracket x^2 \rrbracket_2 &= 3 \cdot 2 + 2 \cdot 1 + 1 = 9 \equiv 4 \pmod{5}, \\
\text{User 3:} \, \llbracket x^2 \rrbracket_3 &= 3 \cdot 2 + 2 \cdot 1 + 1 = 9 \equiv 4 \pmod{5}.
\end{align*}

Next, each user computes the masked values required for secure multiplication of \(x \cdot x^2\):
\begin{align*}
\text{User 1:} \, & x_1 - \llbracket a^2 \rrbracket_1 = 1 - 4 = -3 \equiv 2 \pmod{5}, \\
                     & \llbracket x^2 \rrbracket_1 - \llbracket b^2 \rrbracket_1 = 0 - 0 = 0 \pmod{5}, \\
\text{User 2:} \, & x_2 - \llbracket a^2 \rrbracket_2 = -1 - 3 = -4 \equiv 1 \pmod{5}, \\
                     & \llbracket x^2 \rrbracket_2 - \llbracket b^2 \rrbracket_2 = 4 - 1 = 3 \pmod{5}, \\
\text{User 3:} \, & x_3 - \llbracket a^2 \rrbracket_3 = 1 - 1 = 0 \pmod{5}, \\
                     & \llbracket x^2 \rrbracket_3 - \llbracket b^2 \rrbracket_3 = 4 - 4 = 0 \pmod{5}.
\end{align*}

Each user then transmits the above masked values to Server. The server aggregates the results to reconstruct the global masked values:
\begin{align*}
x - a^2 &= \sum_{i=1}^{3} (x_i - \llbracket a^2 \rrbracket_i) = 2 + 1 + 0 = 3 \pmod{5}, \\
x^2 - b^2 &= \sum_{i=1}^{3} (\llbracket x^2 \rrbracket_i - \llbracket b^2 \rrbracket_i) = 0 + 3 + 0 = 3 \pmod{5}.
\end{align*}

Server then broadcasts these aggregated values \((x - a^2)\) and \((x^2 - b^2)\) to all users to complete the secure multiplication step for computing \(x^3\).
\end{itemize}

\textbf{Global computation:} After completing the two subrounds, each user proceeds to compute a share of the final majority vote polynomial \( F(x) = 2x^3 + 4x \pmod{5} \). Using the broadcast values \((x - a^2)\) and \((x^2 - b^2)\) from Server, and their local input \(x_i\), each user locally evaluates the masked cubic term \(\llbracket x^3 \rrbracket_i\) as follows:
\[
\llbracket x^3 \rrbracket_i \!=\! (x \!-\! a^2)(x^2 \!-\! b^2) \!+\! \llbracket a^2 \rrbracket_i(x^2 \!-\! b^2) \!+\! \llbracket b^2 \rrbracket_i(x \!-\! a^2) \!+\! \llbracket c^2 \rrbracket_i \!\!\!\pmod{5}.
\]

The individual computations are given below:
\begin{align*}
\text{User 1:} \, \llbracket x^3 \rrbracket_1 &= 3 \cdot 1 + 4 \cdot 1 + 0 \cdot 3 + 1 = 8 \equiv 3 \pmod{5}, \\
\text{User 2:} \, \llbracket x^3 \rrbracket_2 &= 3 \cdot 1 + 1 \cdot 3 + 2 = 8 \equiv 3 \pmod{5}, \\
\text{User 3:} \, \llbracket x^3 \rrbracket_3 &= 1 \cdot 1 + 4 \cdot 3 + 2 = 15 \equiv 0 \pmod{5}.
\end{align*}

Each user then substitutes the locally computed values into the final polynomial:
\[
\llbracket F(x) \rrbracket_i = 2 \llbracket x^3 \rrbracket_i + 4x_i \pmod{5}.
\]

The share computations are as follows:
\begin{align*}
\text{User 1:} \, \llbracket F(x) \rrbracket_1 &= 2 \cdot 3 + 4 \cdot 1 = 10 \equiv 0 \pmod{5}, \\
\text{User 2:} \, \llbracket F(x) \rrbracket_2 &= 2 \cdot 3 + 4 \cdot (-1) = 2 \pmod{5}, \\
\text{User 3:} \, \llbracket F(x) \rrbracket_3 &= 2 \cdot 0 + 4 \cdot 1 = 4 \pmod{5}.
\end{align*}

Each user sends their computed share \(\llbracket F(x) \rrbracket_i\) to Server. The server aggregates the values to obtain the final result:
\begin{align*}
F(x) &= \sum_{i=1}^{3} \llbracket F(x) \rrbracket_i 
= \sum_{i=1}^{3} \left(2\llbracket x^3 \rrbracket_i + 4x_i\right) \pmod{5} \\
&= 0 + 2 + 4 = 6 \equiv 1 \pmod{5}.
\end{align*}

This demonstrates that the majority vote polynomial \( F(x) \) can be securely computed via Beaver triples without revealing any individual user's input, while producing an output equivalent to that of the standard non-secure majority voting protocol.

% ==============================================================================================
\begin{comment}
\section{Precomputed Table of Majority Vote Polynomials \( F(\xv) \)} \label{appendix:Precomputed polynomial}

The majority vote polynomial \( F(\xv) \) can be efficiently precomputed once the number of users \( n \) and the tie-breaking policy are determined in the offline phase. Specifically, for each given \( n \) and the tie-breaking rule, the corresponding polynomial can be systematically derived using Eq.~\ref{eq2}.

Table~\ref{table:majority_vote} presents representative examples of precomputed polynomials according to tie-breaking policies. 

\begin{table}[!htb]
    \centering
    %\small
    \setlength{\tabcolsep}{2.5pt}
    \caption{Precomputed majority vote polynomials \( F(\xv) \) according to tie-breaking policies}
    \label{table:majority_vote}
    \begin{tabular}{|c|c|c|}
    \hline
    \#Users & \( \text{sign}(0)\in \{-1, +1\} \) & \( \text{sign}(0) = 0 \) \\ \hline 
    $n=2$ & \( \xv^2 + 2\xv + 2 \pmod{3} \) & \(2\xv \pmod{3} \) \\ 
    $n=3$ & \( 2\xv^3 + 4\xv \pmod{5} \) & \( 2\xv^3 + 4\xv \pmod{5} \) \\
    $n=4$ & \( \xv^4 + 3\xv^3 + \xv + 4 \pmod{5} \) & \( 3\xv^3 + \xv \pmod{5} \) \\
    $n=5$ & \( 3\xv^5 + 2\xv^3 + 3\xv \pmod{7} \) & \( 3\xv^5 + 2\xv^3 + 3\xv \pmod{7} \) \\
    $n=6$ & \( \xv^6 + 4\xv^5 + 5\xv^3 + 4\xv + 6 \pmod{7} \) & \( 4\xv^5 + 5\xv^3 + 4\xv \pmod{7} \) \\ \hline
    \end{tabular}
\end{table}
\end{comment}
% ===========================================================================
\section{Proof of Theorem~\ref{thm:hier_sigsgd}}
\label{appendix:hierarchical_proof}

We establish Theorem~\ref{thm:hier_sigsgd} by presenting a rigorous convergence analysis of the hierarchical \textsc{SignSGD} update rule, following the structure of \cite{Ber18}. We begin by stating the assumptions required for the proof.

%This appendix provides a rigorous convergence analysis of the hierarchical \textsc{SignSGD} update rule, following the structure of \cite{Ber18}. 

\vspace{0.1cm}
\noindent{\em Assumption 1 (Lower bound).} 
$f(\thetav)\ge f^\star$ for all $\thetav$.

\vspace{0.1cm}
\noindent{\em Assumption 2 (Coordinate-wise $L$-smoothness).}
For all $\thetav,\phiv\in\mathbb{R}^d$,
\[
|f(\phiv)-f(\thetav)-\nabla f(\thetav)^\top(\phiv-\thetav)|
\le \frac12 \sum_{j=1}^d L_j(\phiv_j-\thetav_j)^2 .
\]

\vspace{0.1cm}
\noindent{\em Assumption 3 (Unbiased stochastic gradients).}
\[
\mathbb{E}[\nabla\tilde f_j(\thetav)]
= \nabla f_j(\thetav),\quad
\mathbb{E}[(\nabla\tilde f_j(\thetav)-\nabla f_j(\thetav))^2]
\le\sigma_j^2.
\]

\vspace{0.1cm}
\noindent{\em Assumption 4 (Symmetric, unimodal noise).}
Each coordinate of $\nabla\tilde f(\thetav)$ is symmetric and unimodal around its mean.

\vspace{0.2cm}
Based on these assumptions, we now prove the proof of Theorem~\ref{thm:hier_sigsgd}. Users are partitioned into $\ell$ subgroups, each of size $n_1=n/\ell$. Each subgroup produces an unbiased majority estimate that is correct with probability $q>1/2$. The update rule $\thetav_{k+1}=\thetav_k-\eta\sv_k$, together with Assumption~2, yields the descent inequality
\[
f(\thetav_{k+1})
\le f(\thetav_k) - \eta\,\gv_k^{\!\top}\sv_k
+ \frac{\eta^2}{2}\|\vec L\|_1.
\]
Taking expectations,
\begin{equation}
\label{eq:descent_base_final}
\mathbb{E}[f(\thetav_{k+1})]
\le \mathbb{E}[f(\thetav_k)]
- \eta\,\mathbb{E}[\gv_k^{\!\top}\sv_k]
+ \frac{\eta^2}{2}\|\vec L\|_1 .
\end{equation}

For each coordinate $j$, defining the sign error probability 
$\varepsilon_{k,j}=\Pr[s_{k,j}\neq\operatorname{sign}(g_{k,j})]$, one has
\[
\mathbb{E}[g_{k,j}s_{k,j}]
= |g_{k,j}|(1-2\varepsilon_{k,j}),
\]
and therefore
\begin{equation}
\label{eq:inner_product_bound_final}
\mathbb{E}[\gv_k^{\!\top}\sv_k]
= \mathbb{E}[\|\gv_k\|_1]
- 2\sum_{j=1}^d \varepsilon_{k,j}\,\mathbb{E}[|g_{k,j}|].
\end{equation}
The remaining task is to bound 
$\sum_{j=1}^d \varepsilon_{k,j}\,\mathbb{E}[|g_{k,j}|]$.

\vspace{0.15cm}
\noindent\textit{Subgroup-level behavior.}
By Assumptions~3–4 and \cite[Appendix~D]{Ber18}, a single user's sign error probability satisfies
\[
\Pr[X_i\neq\operatorname{sign}(g_{k,j})]
\le\exp\!\left(-\frac{g_{k,j}^2}{2\sigma_j^2}\right).
\]
For a subgroup of size $n_1$, Hoeffding's inequality implies an exponentially decaying probability of subgroup-level majority error:
\[
\Pr[\hat g^{(r)}_{k,j}\neq\operatorname{sign}(g_{k,j})]
\le e^{-c_1 n_1},
\]
for some $c_1>0$ depending on the per-user margin.  
Following the stochastic-smoothness argument in \cite{Ber18}, the total contribution of subgroup errors satisfies
\begin{equation}
\label{eq:sub_noise_final}
\sum_{j=1}^d
\Pr[\text{subgroup error at }j]\,
\mathbb{E}[|g_{k,j}|]
\;\le\;
\frac{1}{\sqrt{n_1}}\|\vec\sigma\|_1.
\end{equation}

\vspace{0.15cm}
\noindent\textit{Global aggregation behavior.}
Let $Y_r=\hat g^{(r)}_{k,j}\operatorname{sign}(g_{k,j})\in\{-1,+1\}$.  
Since each subgroup is correct with probability $q_r>1/2$, letting $q=\min_r q_r$ yields
\[
\mathbb{E}[Y_r]\ge\nu_{\min}:=2q-1>0.
\]
The global majority fails only when $\sum_{r=1}^{\ell}Y_r\le 0$, and Hoeffding's inequality gives
\[
\Pr[s_{k,j}\neq\operatorname{sign}(g_{k,j})]
\le e^{-c_2\ell},
\quad c_2=\frac{(2q-1)^2}{2}.
\]
Defining $C_{\mathrm{hier}} := \sum_{j=1}^d \mathbb{E}[|g_{k,j}|]$,
the weighted global error term is therefore bounded by
\begin{equation}
\label{eq:global_noise_final}
\sum_{j=1}^d
\Pr[\text{global majority error at }j]\,
\mathbb{E}[|g_{k,j}|]
\;\le\;
C_{\mathrm{hier}} e^{-c_2\ell}.
\end{equation}

\vspace{0.15cm}
\noindent\textit{Combined bound.}
Combining \eqref{eq:sub_noise_final} and \eqref{eq:global_noise_final},
\begin{equation}
\label{eq:combined_noise_final}
\sum_{j=1}^d \varepsilon_{k,j}\,\mathbb{E}[|g_{k,j}|]
\le
\frac{1}{\sqrt{n_1}}\|\vec\sigma\|_1
+ C_{\mathrm{hier}}e^{-c_2\ell}.
\end{equation}

Substituting \eqref{eq:combined_noise_final} into \eqref{eq:inner_product_bound_final}, and then into \eqref{eq:descent_base_final}, yields
\[
\begin{aligned}
\eta\,\mathbb{E}[\|\gv_k\|_1]
\le\;&
\mathbb{E}[f(\thetav_k)] - \mathbb{E}[f(\thetav_{k+1})] \\
&+ \eta\!\left(
\frac{2}{\sqrt{n_1}}\|\vec{\sigma}\|_1
+ 2C_{\mathrm{hier}} e^{-c_2\ell}
\right)
+ \frac{\eta^2}{2}\|\vec{L}\|_1 .
\end{aligned}
\]

Summing this inequality over $k=0,\dots,K-1$ and using $f(\thetav_K)\ge f^\star$ gives
\[
\begin{aligned}
\eta\sum_{k=0}^{K-1}\mathbb{E}[\|\gv_k\|_1]
\le\;&
f_0-f^\star
+ K\eta\!\left(
\frac{2\|\vec\sigma\|_1}{\sqrt{n_1}}
+ 2C_{\mathrm{hier}}e^{-c_2\ell}
\right) \\
&\qquad + \frac{K\eta^2}{2}\|\vec L\|_1 .
\end{aligned}
\]

Choosing $\eta=1/\sqrt{K\|\vec L\|_1}$ and defining $N_t=K^2$ leads to
\[
\frac{1}{K}\sum_{k=0}^{K-1}\mathbb{E}[\|\gv_k\|_1]
\le
\frac{\sqrt{\|\vec L\|_1}}{N_t^{1/4}}
\big(f_0-f^\star+\tfrac12\big)
+ \frac{2\|\vec\sigma\|_1}{\sqrt{n_1}}
+ C_{\mathrm{hier}} e^{-c_2\ell}.
\]

Finally, squaring both sides,
\[
\begin{aligned}
\mathbb{E}\!\left[\frac{1}{K}\sum_{k=0}^{K-1}\|\gv_k\|_1\right]^2
\le\;&
\frac{1}{\sqrt{N_t}}
\Big(
\sqrt{\|\vec{L}\|_1}(f_0 - f^\star + \tfrac12)
\\
&\qquad\qquad
+\, \tfrac{2}{\sqrt{n_1}}\|\vec{\sigma}\|_1
+\, C_{\mathrm{hier}} e^{-c_2 \ell}
\Big)^2 .
\end{aligned}
\]

This completes the proof of Theorem~\ref{thm:hier_sigsgd}.
\qed

\section{Proof of Theorem~\ref{thm:hierarchical_secure}}
\label{appendix:security_proofs}

\noindent\textit{Preliminaries.}
%\subsection{Preliminaries}
Let $\mathbb{F}_{p_1}$ and $\mathbb{F}_{p_2}$ be prime fields with $p_1 > n_1$ and $p_2 > \ell$, respectively.  For each subgroup $\mathcal{G}_j$, define $\xv_j = \sum_{i=1}^{n_1}\xv_{i,j}$, $\sv_j = \mathrm{sign}(\xv_j)$, and let $F(\cdot)$ be the finite-field majority polynomial satisfying $F(\xv_j) \!=\! \sv_j \pmod{p_1}$.  All multiplications are carried out using Beaver triples $(\av,\bv,\cv)$ with $\cv \!=\! \av\!\cdot \!\bv$, generated in the offline phase and sampled independently of all inputs.

We use additive secret sharing: for a secret $\zv$, parties hold shares $\{\llbracket \zv\rrbracket_i\}$ such that $\sum_i \llbracket \zv\rrbracket_i = \zv$.  For inputs $\xv,\yv$ and masks
$\av,\bv$, the protocol opens $\deltav = \xv - \av$, $\epsilonv = \yv - \bv$, which are publicly revealed.  By construction of the triples, their distribution is dominated by the unknown random share of at least one honest user in the global system.

\noindent\textit{Proof of Theorem~\ref{thm:hierarchical_secure}.}
Let $\mathcal{A}$ corrupt a set $\mathcal{C}\subseteq[n]$ of at most $t\le n-1$ users. The simulator $\mathsf{SIM}$ is given the corrupted inputs $\{\xv_{i,j}\}_{i\in\mathcal{C}}$ and the leakage $\{\sv_j\}_{j=1}^{\ell}$ and $\sv$.

\begin{enumerate}
    \item \textit{Preprocessing.}  
    $\mathsf{SIM}$ samples all Beaver triples $(\av,\bv,\cv)$ gate-wise over $\mathbb{F}_{p_1}$ and $\mathbb{F}_{p_2}$ exactly as in the real preprocessing, and distributes additive shares to all parties (including corrupted ones).

    \item \textit{Intra-subgroup simulation.}  
    For each subgroup $\mathcal{G}_j$, the simulator invokes Lemma~\ref{lem:intra_simulation} with the given output $\sv_j$ to generate a simulated transcript of the local computation of $F(\xv_j)$, including all masked openings and the final reconstruction of $\sv_j$.

    \item \textit{Inter-group simulation.}  
    Using $\{\sv_j\}_{j=1}^{\ell}$ and $\sv$, the simulator invokes Lemma~\ref{lem:inter_simulation} to simulate the inter-group aggregation and the reconstruction of the global majority $\sv$.\!\!
\end{enumerate}

By Lemma~\ref{lem:local_privacy} (stated later in this section), all public masked openings in both layers are independent of the honest inputs and can be replaced by uniform samples.  By Lemma~\ref{lem:intra_simulation}, the intra-subgroup transcripts are simulatable given $\sv_j$, and by Lemma~\ref{lem:inter_simulation}, the inter-group transcript is simulatable given $\{\sv_j\}$ and $\sv$.

Therefore,
\[
\mathsf{REAL}_{\mathcal{A}}\big(\{\xv_{i,j}\}_{i\in\mathcal{C}}\big)
\;\approx_c\;
\mathsf{SIM}_{\mathcal{A}}\big(\{\xv_{i,j}\}_{i\in\mathcal{C}},\,
\{\sv_j\}_{j=1}^{\ell},\,\sv\big),
\]
which proves Theorem~\ref{thm:hierarchical_secure}.
\qed

% --------------------- lemma 이동
%\subsection{Lemma 1 (Privacy of Beaver Masked Openings)}
\begin{lemma} [Privacy of Beaver Masked Openings]
\label{lem:local_privacy}

\textbf{\\Claim.}
Suppose at most $t\le n-1$ users are corrupted globally.  Then, for every multiplication gate in the protocol, the publicly opened pair $(\deltav,\epsilonv) = (\xv-\av,\yv-\bv)$ is computationally indistinguishable from a pair of uniform and input-independent elements of $\mathbb{F}_{p_1}$ (or $\mathbb{F}_{p_2}$ at the inter-group layer).

\begin{proof}
In the offline phase, all $n$ users jointly generate Beaver triples $(\av,\bv,\cv)$ via an MPC protocol, where $\av$ and $\bv$ are sampled uniformly and independently of the online inputs $(\xv,\yv)$.  Since at most $t\le n-1$ users are corrupted, there exists at least one honest user whose local randomness contributes an unknown, uniformly random share to $\av$ and $\bv$.  From the adversary’s point of view, the masks $\av,\bv$ are therefore distributed as fixed offsets plus independent uniform elements of $\mathbb{F}_{p_1}$.
Consequently, $\deltav = \xv-\av, \epsilonv = \yv-\bv$ are distributed as fixed (adversary-dependent) offsets plus independent uniform field elements, and hence are indistinguishable from uniform and independent of $(\xv,\yv)$. The same argument applies at the inter-group layer over $\mathbb{F}_{p_2}$.
\end{proof}
\end{lemma}

%\subsection{Lemma 2 (Simulatability of Intra-Subgroup Computation)}
\begin{lemma} [Simulatability of Intra-Subgroup Computation]
\label{lem:intra_simulation}

\textbf{\\Claim.}
Fix a subgroup $\mathcal{G}_j$ and its majority output $\sv_j = F(\xv_j)$.  Given $\sv_j$, the complete transcript of the computation of $F(\xv_j)$ inside $\mathcal{G}_j$--including all masked openings and the final reconstruction of $\sv_j$--is simulatable without knowing any honest user inputs.

\begin{proof}
The intra-subgroup computation of $F(\xv_j)$ consists of additions on secret shares and Beaver-based multiplications over $\mathbb{F}_{p_1}$. By Lemma~\ref{lem:local_privacy}, the masked openings $(\deltav,\epsilonv)$ at each multiplication gate are uniform and input-independent, and thus can be simulated by sampling uniform elements in $\mathbb{F}_{p_1}^2$.
Given the target output $\sv_j$, the simulator first samples additive shares $\{\llbracket \sv_j\rrbracket_i\}_{i\in\mathcal{G}_j}$ uniformly at random subject to $\sum_{i\in\mathcal{G}_j} \llbracket \sv_j\rrbracket_i = \sv_j$. It then simulates the finite-field arithmetic operations step-by-step: for each multiplication gate, it samples uniform masked openings $(\hat{\deltav},\hat{\epsilonv})$, updates the simulated shares according to the Beaver correctness relation, and ensures that the final reconstruction of the output equals $\sv_j$.
Because all public openings are independent of the true inputs and the only plaintext value revealed is $\sv_j$, which the simulator is given, the resulting simulated transcript is computationally indistinguishable from the real execution for subgroup $\mathcal{G}_j$.
\end{proof}
\end{lemma}

%\subsection{Lemma 3 (Simulatability of Inter-Group Aggregation)}
\begin{lemma} [Simulatability of Inter-Group Aggregation]
\label{lem:inter_simulation}

\textbf{Claim.}
Given the subgroup majority outputs $\{\sv_j\}_{j=1}^{\ell}$ and the final global majority $\sv = \mathrm{sign}(\sum_{j=1}^{\ell}\sv_j)$, the entire inter-group transcript is simulatable without access to any honest user inputs.

\begin{proof}
At the inter-group layer, the inputs to the finite-field arithmetic operations are the plaintext subgroup outputs $\{\sv_j\}_{j=1}^{\ell}$ (reconstructed from their shares) or, equivalently, their secret sharings over $\mathbb{F}_{p_2}$.  Given $\{\sv_j\}$ and $\sv$, the simulator can secret-share each $\sv_j$ among the parties by sampling shares uniformly at random subject to summing to $\sv_j$. It then simulates the inter-group finite-field arithmetic operations step-by-step: for each multiplication, it samples uniform masked openings $(\hat{\delta},\hat{\epsilon})$ in $\mathbb{F}_{p_2}^2$ and updates the shares accordingly so that the final reconstruction equals the given global majority $\sv$.
By Lemma~\ref{lem:local_privacy}, the real masked openings are input-independent uniforms, so replacing them with uniform samples preserves the distribution.  Since the simulator exactly reproduces the observable subgroup outputs $\{\sv_j\}$ and the final output $\sv$, the simulated inter-group transcript is indistinguishable from the real one.
\end{proof}
\end{lemma}

\ifCLASSOPTIONcaptionsoff
  \newpage
\fi

\bibliographystyle{IEEEtran}
\bibliography{reference}

\end{document}

%% file: macros.tex
\long\def\comment#1{}

\newfont{\bbb}{msbm10 scaled 700}

\newfont{\bb}{msbm10 scaled 1100}

\newcommand{\av}{{\bf a}}
\newcommand{\bv}{{\bf b}}
\newcommand{\cv}{{\bf c}}

\newcommand{\gv}{{\bf g}}

\newcommand{\mv}{{\bf m}}

\newcommand{\sv}{{\bf s}}

\newcommand{\xv}{{\bf x}}
\newcommand{\yv}{{\bf y}}
\newcommand{\zv}{{\bf z}}

\newcommand{\deltav}{\hbox{\boldmath$\delta$}}

\newcommand{\epsilonv}{\hbox{\boldmath$\epsilon$}}

\newcommand{\phiv}{\hbox{\boldmath$\phi$}}

\newcommand{\thetav}{\hbox{\boldmath$\theta$}}

\renewcommand{\vec}{{\rm vec}}

\newcommand{\RED}{\color[rgb]{1.00,0.10,0.10}}